\documentclass[fleqn,10pt]{wlscirep}
\usepackage[utf8]{inputenc}
\usepackage[T1]{fontenc}
\usepackage{lineno}
\usepackage{tikz}
\usetikzlibrary{decorations.pathreplacing, decorations.pathmorphing, positioning, arrows, fit, backgrounds}

\title{Estimating Global Input Relevance and Enforcing Sparse Representations with a Scalable Spectral Neural Network Approach}

\author[1,*]{Lorenzo Chicchi}
\author[1]{Lorenzo Buffoni}
\author[1]{Diego Febbe}
\author[1]{Lorenzo Giambagli}
\author[1]{Raffaele Marino}
\author[1]{Duccio Fanelli}
\affil[1]{Department of Physics and Astronomy, University of Florence, Sesto Fiorentino, Italy \\ INFN, Italy}

\affil[*]{lorenzo.chicchi@unifi.it}

\begin{abstract}
In machine learning practice it is often useful to identify relevant input features. Isolating key input elements, ranked according their respective degree of relevance, can help to elaborate on the process of decision making. Here, we propose a novel method to estimate the relative importance of the input components for a Deep Neural Network. This is achieved by leveraging on a spectral re-parametrization of the optimization process. Eigenvalues associated to input nodes provide in fact a robust proxy to gauge the relevance of the supplied entry features. Notably, the spectral  features  ranking is performed automatically, as a byproduct of the network training, with no additional processing to be carried out. Moreover, by leveraging on the regularization of the eigenvalues, it is possible to enforce solutions making use of a minimum subset of the input components, increasing the explainability of the model and providing sparse input representations. The technique is compared to the most common methods in the literature and is successfully challenged against both synthetic and real data.
\end{abstract}

\begin{document}

\flushbottom

\maketitle

%
%
\section{Introduction}
Deep Neural networks (DNNs) \cite{bishop2023deep,goodfellow2016deep,lecun2015deep, prince2023understanding,carleo2019machine} are among the most popular machine learning tools, and, as such, routinely employed in a vast plethora of applications \cite{reiser2022graph,chmiela2017machine, chicchi2023frontiers, biancalani2021deep, baldi2014searching}. As for other machine learning techniques, and when employed for e.g. classification and/or regression tasks, DNNs can automatically extract valuable information from the supplied input vector. This amounts to isolate crucial components to implement unmanned decision making strategies. In many cases of interest, input data are multidimensional. In general, it is hence difficult to selectively identify the subset of attributes which prove crucial for the analysis. Moreover, it is also challenging to determine whether the model’s decisions are based on a large portion of the input components or just a minimal subset. To identify key input features could be indeed relevant for (i) devising apt solutions for data compression and compactification; (ii)  providing critical insights into the processes that led to the produced output.

Numerous methods have been proposed to shed light on the pathways that form the basis of decision of a trained model \cite{zhang2021survey,dwivedi2023explainable,shrikumar2017learning,swartout1993explanation,montavon2018methods,baehrens2010explain}. Some examples include - but are not limited to - LRP ( Layer-Wise Relevance Propagation) \cite{bach2015pixel, montavon2019layer, bohle2019layer, yang2018explaining}, SHAP (SHapley Additive exPlanations)\cite{lundberg2017unified}, IG (Integrated Gradients) \cite{sundararajan2017axiomatic} and LIME (Local Interpretable Model-Agnostic Explanations) \cite{lime}. The  Layer-Wise Relevance Propagation method backprogates the relevance, a quantity that has multiple definition depending on different rules, from the output to the input space, to reach the relevant features that are provided at the entry layer. SHAP is a game theoretic approach based on the classic Shapley values, which connects optimal credit allocation to local output explanation. IG is a gradient-based approach and, finally, LIME supplies information about features relevance by approximating the model with a local, interpretable model.  

To the best of our knowledge, the vast majority of  existing methods are however input-dependent. As such, they require a certain number of operations to be carried out after training, for each examined input, so as to assess the relevance of the features for any given supplied item. For most of the aforementioned methods, estimating the global relevance of the input components requires additional computation, typically involving averaging over the entire dataset. More importantly, none of these methods provides a straightforward way to guide the model during training to reach a solution that relies on either a large or a small subset of input components, as they operate on post-training models and play no role during the training phase. This latter property could be especially valuable when deep neural networks are applied to regression tasks in various scientific domains \cite{belfiore2025machine, novakovsky2023obtaining, chicchi2023frontiers, ginolfi2025inferring, camaiani2022model}. In fact, in some research fields, understanding which experimental observations are relevant—and how they are used by the model to solve the task—is sometimes more important than having an opaque model that makes accurate predictions. Identifying a minimal set of input components necessary for solving the task and forcing the model to operate on that sparse representation can help uncover natural relationships among those components.

Here, we put forward a different method which yields a global metric to assess the relevance of input features to a general feedforward neural network. The procedure applies to the whole dataset  and not just to individually examined items. Furthermore, the evaluation of the features is obtained as a byproduct of the training with no additional processing to be carried out. At the end of the training a ranking of the supplied feature is returned, as determined by a set of scalar parameters (one for each distinct feature provided as an input). As mentioned, and unlike the aforementioned methods, the proposed metric is not input dependent. At variance, it provides a global insight on how the model oversees the input space, upon learning. 

It is important to emphasize that the method we propose is not intended to outperform existing techniques in the literature, some of which remain preferable, especially when the goal is to understand the decision-making process for a specific input. However, the proposed algorithm differs fundamentally from traditional approaches. In fact, it relies on parameters that play a crucial role both during the training phase and the successive process aimed at assessing feature relevance. The integration of these two aspects within a single framework equips the system with the inherent ability to steer the model toward solutions that rely on a reduced number of input components, thereby identifying a minimal subset of features sufficient to efficiently solve the task. As already stated, this capability can be extremely valuable in contexts where explainability is critical (e.g., science, medicine), and where enforcing explainable solutions from the training phase may be the preferred approach.

To reach this goal, we will make use of a recently introduced scheme for handling the optimization of feedforward networks, known as \textit{spectral parametrization} \cite{giambagli2021machine, chicchi2021training,chicchi2023recurrent, chicchi2024complex, buffoni2022spectral, giambagli2024student}. The mathematical foundation of the spectral methodology lies within the theory of networks. From this angle, feedforward neural networks can be pictured via a dedicated spectral decomposition  that materializes, in direct space, in a collection of adjacency matrices for the linear mapping of the signal across adjacent layers. As we shall argue, the optimized eigenvalues, as made accessible at the end of the training, will return an automatic and self-consistent ranking of the individual input features.  

The paper is organized as follows. In the next Section we will review the mathematical basis of the spectral approach to neural network training and introduce the propose ranking strategy. Then, we will turn to discussing a first application to independent and correlated Gaussian data. These are mock data, with different level of complexity, to be regressed by the spectral neural network. In the subsequent Section the proposed method is challenged against the celebrated MNIST database, operated in different modalities. As we will show, the spectral based feature detection method is able to isolate a subset of relevant pixels that prove central for image classification. Then, we turn to challenging the method against an additional dataset, made of stellar spectra. The goal here is to obtain information on relevant frequency bands for the assessment of stellar metallicity. A section is then dedicated to compare our method with some of the most commonly used techniques for inferring input feature relevance. As we will show, the spectral scores are in good agreement with those obtained using the other techniques considered. Finally, we sum up and draw our conclusions.

\subsection{On the spectral parametrization}

Let us begin by reviewing the basic of the spectral method \cite{giambagli2021machine}. We denote with $l-1$ and $l$ two consecutive layers of a feedforward neural network. Assume the layers to be respectively composed by $N$ and $M$ nodes. The vector $\textbf{x}^{l-1}=\left(x^{l-1}_1,\hdots,x^{l-1}_N\right)$ (resp. $\textbf{x}^{l}=\left(x^{l}_1,\hdots,x^{l}_M\right)$) stores information on the activity of the nodes belonging to layer $l-1$ (resp. $l$). The linear transfer between the two adjacent layers is operated by a rectangular $M\times N$ matrix $W$, whose elements read $w_{ij}$. Hence,  $\textbf{x}^l = f(W\textbf{x}^{l-1})$, where $f$ is a non-linear function. Under the spectral paradigm, the above operation can be recast in an equivalent form that assumes dealing with the bipartite direct network made of the $N+M$ nodes involved in the transfer. The state vector at the departure layer can be written as $\textbf{z}^{l-1} = \textbf{x}^{l-1}=\left(x^{l-1}_1, x^{l-1}_2, \hdots x^{l-1}_{N}, 0, 0, \hdots 0 \right)$ where we made explicit that the non trivial activity is solely localized on the nodes of the first layer. Then, $\textbf{z}^{l}= f(A \textbf{z}^{l-1})$, where the last $M$ elements of vector $\textbf{z}^{l}$ refer to the activity on the destination layer; $A$ is a $(M+N) \times (M+N)$ matrix with a $M\times N$ sub-diagonal block that coincides matrix $W$, as illustrated in the Methods section. It should be noted that the diagonal entries of matrix $A$ do not reflect on the produced activity on layer $l$, see Figure \ref{spect_diag}. One can hence invoke  a spectral decomposition of the square transfer matrix, as $A=\Phi \Lambda \Phi^{-1}$ where (i) $\Phi$, the matrix of the eigenvectors, is lower-block triangular with ones on the diagonal, (ii) $\Lambda$ identifies the diagonal matrix formed with the eigenvalues ($\lambda_i$, $i=1, \hdots N+M$) of $A$. Given the specific form  of  matrix $\Phi$, it can be shown \cite{buffoni2022spectral, chicchi2021training} that $\Phi^{-1} = 2\mathbb{I} -\Phi$. In practical terms, the elements of $A$, namely those involved in the self-consistent definition of the relevant transfer matrix $W$, can be written as a function of the eigenvalues $\lambda_i$ and the non trivial entries $\psi_{ij}$ of the eigenvector matrix $\Phi$. An explicit calculation can be performed that yields $ w_{i\rightarrow j} = w_{ji} = (\lambda_i - \lambda_j)\psi_{ji}$, where $i=1,...,N$ refers to nodes located on layer $l-1$ and $j=1,...,M$ to those assigned to layer $l$. Further technical details are provided in the Methods section. A direct correspondence can be hence drawn between individual eigenvalues and nodes within the sampled collection. 
 Training a feedforward network under the spectral standpoint amounts to minimize the loss (a suitable metric that gauges the distance between produced and expected output at the exit layer) with respect to the eigenvalues and eigenvectors. In light of the existing correspondence, the magnitude of the optimized eigenvalues can be used as a reliable marker of the nodes' relevance, an idea that has been explored in \cite{buffoni2022spectral,chicchi2021training,giambagli2024student}. An effective trimming strategy was in particular proposed in \cite{buffoni2022spectral} which enables for the trained network to be considerably shrunk in size by removing nodes flagged as unimportant.  By leveraging on the eigenvalues and their associated magnitudes, it was possible in \cite{giambagli2024student} to estimate the hidden dimension of the target objective function. By expanding further on these ideas, we will hereafter propose an effective input feature detection algorithm. To this end, we chose to operate under the simplifying assumption $\lambda_j=0$ (i.e. by forcing to zero the eigenvalues associated to destination nodes) yielding:
\begin{equation}
w_{ji} = \lambda_i\psi_{ji}.
\label{eq_spectral_riducted}
\end{equation}
The spectral rewriting of the inter-nodes weights $w_{ij}$ yielded therefore a factor, the eigenvalue $\lambda_i$,  which multiplies all the links emanating from the departure node $i$. The larger the magnitude of the optimized $\lambda_i$ the more relevant the associated node $i$ for the processing of the information across the network. Building on this observation, we will show that the trained eigenvalues associated to the starting  layer $l=1$ can be effectively employed to automatically discriminate between irrelevant and relevant input features. More specifically, we will demonstrate,  that eigenvalues (multiplied by the norm of the corresponding eigenvector entry) can be exploited to rank the relative importance of the features otherwise deemed as irrelevant. The method will be tested against classification problems constructed from mock and real datasets. For all examined cases, a 
penalty is added to the loss function as in the spirit of  \cite{giambagli2024student, giambagli2024jstat}. This is a $L_2$ or a $L_1$ regularization of $\lambda_i$ which forces  eigenvalues associated with irrelevant components to be set to zero (and not just left unaltered to their initial values, possibly drowned in the sea of their relevant homologue). It is worth stressing that in  principle it is enough to have just the first layer parameterized as spectral for this analysis to apply. For further details on the training procedure refers to the Methods section.

\begin{figure}[h]
    \centering
    \begin{tikzpicture}[>=stealth]
        \foreach \i in {1,...,5} {
            \node[circle,draw,inner sep=4pt, thick] (h\i) at (-10,-\i) {};
            \fill[blue, opacity=0.5] (h\i) circle (0.12);
        }
        
        \foreach \i in {1,...,5} {
            \foreach \j in {1,...,3} {
                \node[circle,draw,inner sep=4pt, thick] (out\j) at (-8,-\j-1) {};
                \draw[red, ->] (h\i) -- (out\j);
            }
        }
        
        \draw[->, line width=1mm] (-7.5,-3) -- (-6.5,-3);
        
        \foreach \i in {1,...,5} {
            \node[circle,draw,inner sep=4pt, thick] (h2\i) at (-6,-\i) {};
        }
        
        \foreach \i in {1,...,5} {
            \foreach \j in {1,...,3} {
                \node[circle,draw,inner sep=4pt, thick] (out2\j) at (-4,-\j-1) {};
                \draw[red] (h2\i) -- (out2\j);
            }
        }

        \foreach \i in {1,...,3} {
            \fill [violet, opacity=0.5] (out2\i) circle (0.12);
        }

        \node [above=0.5cm] at (h1) {\large$l-1$};
        \node [above=0.5cm] at (out1) {\large$l$};
        \node [above=0.5cm] at (h21) {\large$l-1$};
        \node [above=0.5cm] at (out21) {\large$l$};

        \foreach \i in {1,...,5} {
            \node[circle,draw,inner sep=4pt, thick] (h\i) at (0,-\i) {};
            \fill[blue, opacity=0.5] (h\i) circle (0.12);
        }
        
        \foreach \i in {1,...,5} {
            \foreach \j in {1,...,3} {
                \node[circle,draw,inner sep=4pt, thick] (out\j) at (2,-\j-1) {};
                \draw[red, ->] (h\i) -- (out\j);
            }
        }

        \foreach \i in {1,...,5}{
            \draw[->, green] (h\i) edge [out=45,in=135,loop] (h\i);
        }
        
        \foreach \i in {1,...,3} {
            \draw[->, orange] (out\i) edge [out=45,in=135,loop] (out\i);
        }
        
        \draw[->, line width=1mm] (2.5,-3) -- (3.5,-3);
        
        \foreach \i in {1,...,5} {
            \node[circle,draw,inner sep=4pt, thick] (h2\i) at (4,-\i) {};
            \fill[cyan, opacity=0.5] (h2\i) circle (0.12);
        }
        
        \foreach \i in {1,...,5} {
            \foreach \j in {1,...,3} {
                \node[circle,draw,inner sep=4pt, thick] (out2\j) at (6,-\j-1) {};
                \draw[red] (h2\i) -- (out2\j);
            }
        }

        \foreach \i in {1,...,3} {
            \fill [violet, opacity=0.5] (out2\i) circle (0.12);
        }
        
        \foreach \i in {1,...,3} {
            \draw[->, orange] (out2\i) edge [out=45,in=135,loop] (out2\i);
        }

        \foreach \i in {1,...,5}{
            \draw[->, green] (h2\i) edge [out=45,in=135,loop] (h2\i);
        }

        \node [above=0.7cm] at (h1) {\large$l-1$};
        \node [above=0.7cm] at (out1) {\large$l$};
        \node [above=0.7cm] at (h21) {\large$l-1$};
        \node [above=0.7cm] at (out21) {\large$l$};

    \end{tikzpicture}
    
    \vspace{1cm}
    
    \begin{tikzpicture}

        \tikzset{decoration={snake,amplitude=.4mm,segment length=2mm, post length=0mm,pre length=0mm}}

        \draw[thick] (-10,0) rectangle (-6,4);
        \draw (-10,0) rectangle (-7.5,1.5);
        \fill [red, opacity=0.5] (-10,0) rectangle (-7.5,1.5);


        \draw[thick] (-5.5,0) rectangle (-5,4);
        \fill[opacity=0.5, blue] (-5,4) rectangle (-5.5,1.5);

        \node at (-4.5,1.75) {\huge$=$};


        \draw[thick] (-4,0) rectangle (-3.5,4);
        \fill[opacity=0.5, violet] (-4,0) rectangle (-3.5,1.5);

        \node at (-8,-0.5) {\large$A$};
        \node at (-5.25,-0.5) {\large$\textbf{z}^{l-1}$};
        \node at (-3.85,-0.5) {\large$\textbf{z}^{l}$};
        
        \draw[thick] (0,0) rectangle (4,4);
        \draw (0,0) rectangle (2.5,1.5);
        \fill [red, opacity=0.5] (0,0) rectangle (2.5,1.5);
        \draw[decorate, green, thick] (0,4) -- (2.5,1.5);
        \draw[decorate, orange, thick] (2.5,1.5) -- (4,0);


        \draw[thick] (4.5,0) rectangle (5,4);
        \fill[opacity=0.5, blue] (5,4) rectangle (4.5,1.5);

        \node at (5.5,1.75) {\huge$=$};


        \draw[thick] (6,0) rectangle (6.5,4);
        \fill[opacity=0.5, violet] (6,0) rectangle (6.5,1.5);
        \fill[opacity=0.5, cyan] (6,1.5) rectangle (6.5,4);

        \node at (2,-0.5) {\large$A$};
        \node at (4.85,-0.5) {\large$\textbf{z}^{l-1}$};
        \node at (6.25,-0.5) {\large$\textbf{z}^{l}$};
        
    \end{tikzpicture}

    \caption{The linear information flow for a modified network with the inclusion of diagonal entries (eigenvalues) in the triangular matrix $A$. The eigenvalues introduce self-loops (as depicted on the right), and this is at variance with the usual setting without loops (on the left). Notice that the bottom elements of \(\textbf{z}^l\) are identical in both cases, yielding  exactly the same activation vector upon linear transfer}

\label{spect_diag}
\end{figure}

\section{Results}
\label{sec:res}

The method discussed above for input feature detection is hereafter challenged against different datasets. We will begin in particular by considering a simple multidimensional Gaussian dataset with independent entries. Then, we will turn to considering mock data generated from correlated Gaussian variables. After that, we will move forward to work with the celebrated MNIST dataset and, finally, we will focus on an experimental catalogue of stellar spectra relevant for astrophysical applications. 

\subsection{Independent Gaussian distributions dataset}
\label{sec:independent_gaussian_results}
 
\begin{figure}
    \centering
    \includegraphics[width=0.7\textwidth]{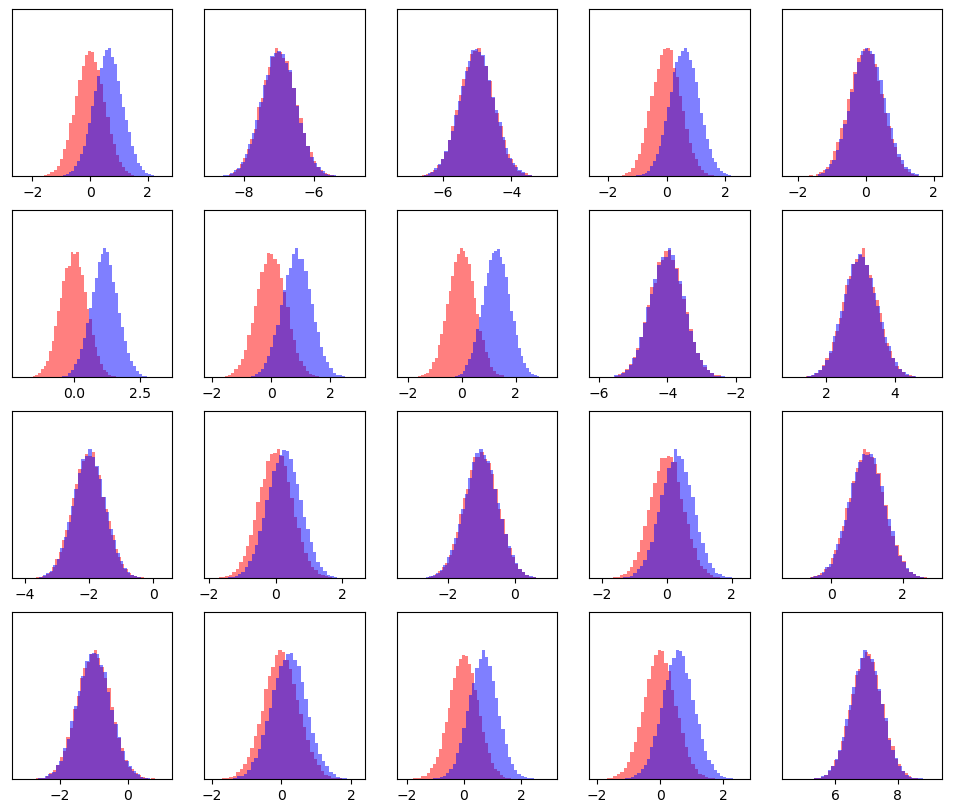}
    \caption{\textbf{Input vector components distributions}. Distribution of the values of the 20 components of the input vectors. The distributions for the data belonging to the first (resp. second) class are represented in red (resp. blue). For some components, the two classes are indistinguishable, while in some other cases the two distributions are only partially overlapping.}
    \label{fig:input_distributions_gauss}
\end{figure}

 The first dataset is structured with 2 target classes and input vectors with 20 independent features. Each data component is sampled from a Gaussian distribution, see Fig. \ref{fig:input_distributions_gauss}. While some components are drawn from distributions which are set to be identical across classes, others display only partial overlap. Components with identical distributions offer no discriminatory information. As such, they prove ineffective for classification purposes. Conversely, the other components are pivotal, the smaller the overlap the larger their relevance in promoting the sought discrimination. For this test, we generated $10^5$ examples.

\begin{figure}
    \centering
    \includegraphics[width=0.9\textwidth]{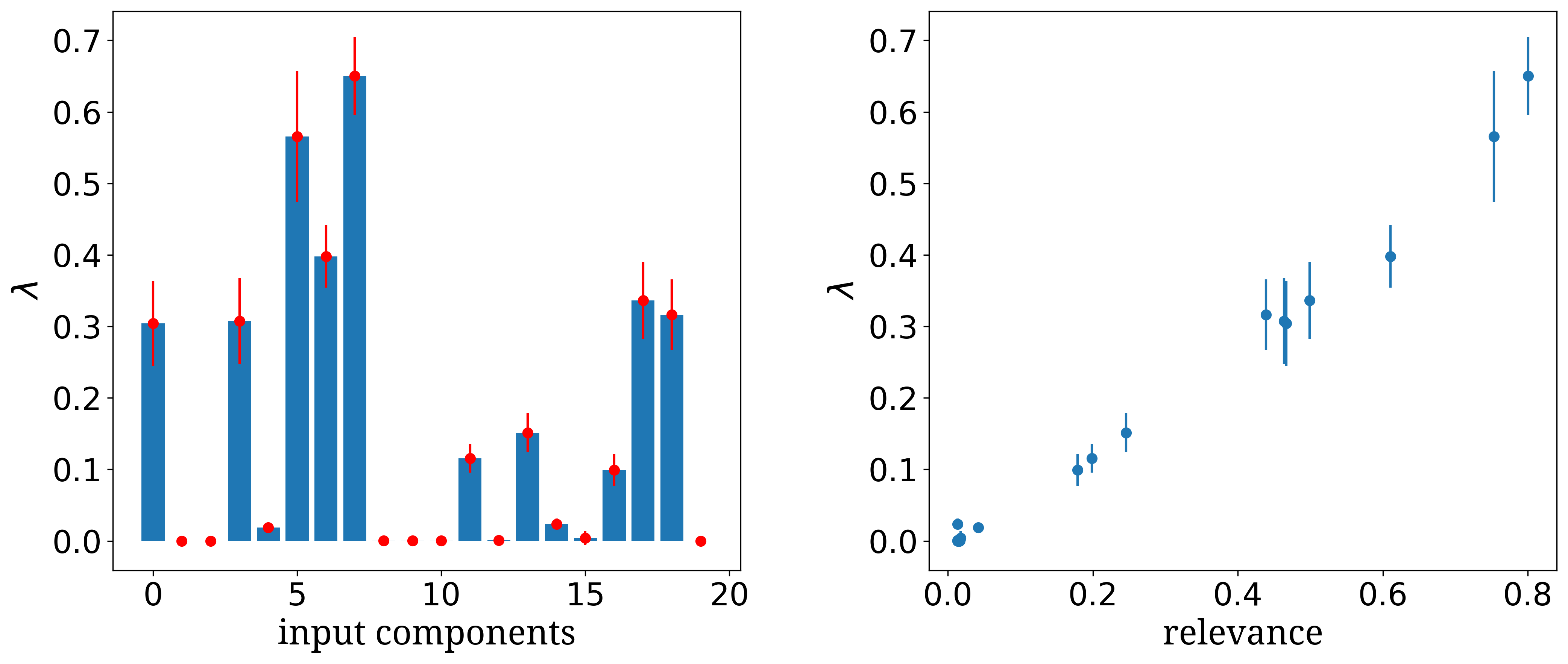}
    \caption{The left panel displays the 20 post-training eigenvalues associated to the input layer, for the \textit{simple Gaussian dataset}. The red dots and the red bars correspond to the average values and the standard deviations of the results obtained by repeating the experiments for 10 independent realizations. In the right panel, the same post-training eigenvalues are reported against a relevance parameter that gauges the distance between the distributions (respectively blue and red, in Fig. \ref{fig:input_distributions_gauss}) from which the components are eventually drawn. The distance is defined as the absolute value of the difference between the two means.}
    \label{fig:multiplot_gauss}
\end{figure}

A spectral feedforward neural network, parametrized as stipulated by  Eq. (\ref{eq_spectral_riducted}), has been trained with the aim of assigning input from the above dataset to the correct class of origin. The employed network consists of three hidden layers with, respectively, 100, 100, and 50 neurons. In addition, the network accommodates for an input layer of size 20 and an output layer of size 2. At the end of the training, we analyzed the 20 eigenvalues, associated with the input layer as in the spirit of the above discussion (with an abuse of language we will from here on call eigenvalues the rescaled version of this latter which incorporate the norm of the eigenvectors, as stipulated in the Methods section). As shown in the left panel of Fig. \ref{fig:multiplot_gauss},  eigenvalues entangled with irrelevant features are all zero, while eigenvalues which can be traced back to relevant input components are non-zero. Hence, the eigenvalues associated to the input layer can efficiently tag features identified as relevant by the network itself to solve the assigned task. The magnitude of non-zero eigenvalues, i.e., those associated with relevant components, is found to correlate with a direct measure of the relative distance between the distributions from which individual components are drawn 
(see the right panel of Fig. \ref{fig:multiplot_gauss}). Since the distributions differ only in their mean value, we quantify the above distance as the absolute value of the difference between the two means. The Pearson correlation ($\rho$) between post-training eigenvalues and the distance between distributions was measured over ten independent training runs, yielding $\rho= 0.981 \pm 0.008$. This strong correlation provides compelling evidence against the presence of spurious relationships between the two quantities. Given the linear nature of the analyzed dataset, we additionally solved the task using logistic regression, allowing us to compare our results with the statistical measures derived from this model. In particular, we compared the post-training eigenvalues with the coefficients obtained through LASSO regularization, which, in this setup, can serve as an alternative benchmark. We observed a strong agreement between the eigenvalues and the LASSO coefficients, with a very high correlation ($\rho = 0.983 \pm 0.007$).


\subsection{Correlated Gaussian distributions dataset}

 Unlike the dataset just analyzed, real datasets often exhibit correlations between input vector components, and crucial task-solving information are frequently concealed within those correlations. With this motivation in mind, we tested the proposed spectral feature detection method on a different dataset that contains pairwise correlations between input vector components. The dataset is still organized into two distinct classes that are to be eventually identified.  Input vectors are now made of 10 components. The first 8 components are defined as follows:
\begin{equation}
\begin{split}
    &x_{2n}  \sim \mathcal{N}(0,1),\\
    &x_{2n+1} = x_{2n}  \;\;\;\;\;\;\;\;\;\;\;\;\;\;\;\;\;\;\;\;\;\;\;\;\;\;\;\;\;\;\;\;\;\;\;\; \text{if} \;\; c(\vec{x}) = 0,\\
    &x_{2n+1} =\begin{cases}
                x_{2n}\; \text{with prob.}\; 1-p_n\\
                -x_{2n}\; \text{with prob.}\; p_n
               \end{cases}  \;\; \text{if} \;\; c(\vec{x}) = 1,
\end{split}
\end{equation}
where $c(\vec{x})$ is the target class of the input and the index $n\in  \{0,1,2,3\}$ denotes the four pairs. The last two components are independently drawn from two identical normal distributions $\mathcal{N}(0,1)$ regardless of the class of belonging:
\begin{equation}
        x_8, x_9 \sim \mathcal{N}(0,1).
\end{equation}

\begin{figure}
    \centering
    \includegraphics[width=0.8\textwidth]{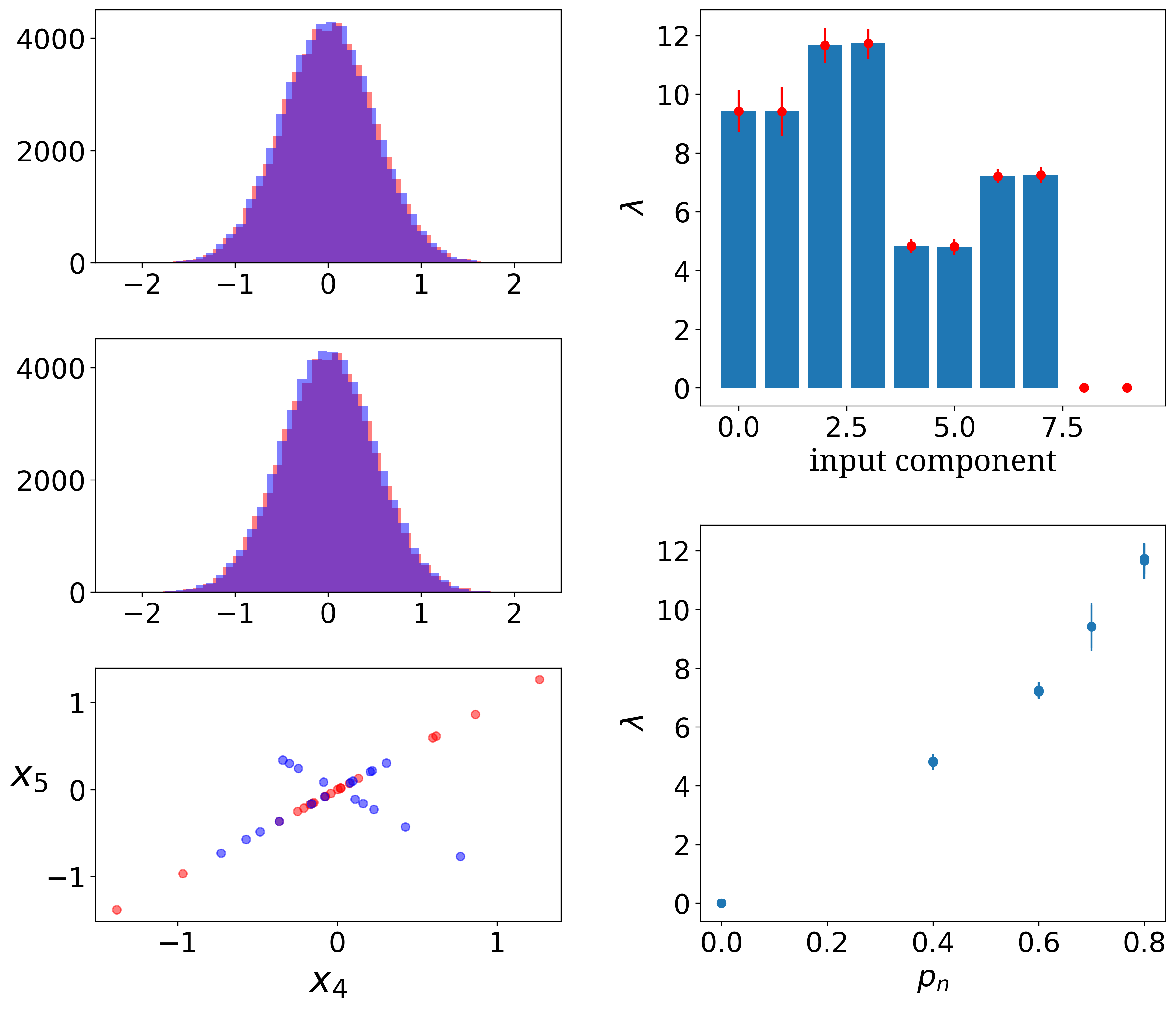}
    \caption{Left column: the distributions of the values of the fourth and fifth components of the input vectors in the \textit{correlated Gaussian dataset} are shown in the first two panels. In the bottom-left panel, the fifth component is plotted against the corresponding fourth component for some input vectors of the same dataset. Right column: the upper-right panel reports the ten post-training eigenvalues (refereed to the input layer) for the \textit{correlated Gaussian dataset}. Red dots and red bars quantifies the recorded average values and the associated standard deviations, as obtained by averaging over 10 independent realizations. In the bottom-right panel the same post-training eigenvalues are plotted against the parameter $p_n$, as defined in the main text.}
    \label{fig:corr_multiplot_resuts}
\end{figure}

As a first noticeable observation, we remark that it is not possible to separate the two datasets by solely relying on independent observations of the respective vector components.  Indeed, crucial information are reflected in how pairs of components are mutually correlated (see left panels of Fig. \ref{fig:corr_multiplot_resuts}).

To better understand the relationship between the different components, let us focus on the first two entries of the input vectors, $x_0$ and $x_1$. For  data belonging to class $0$ (as we label the first of the two) $x_0 = x_1$, that is, the first two components are perfectly correlated. Conversely, the two components of data belonging to the other class  (denoted by $1$) are equal with probability $1-p_0$ and have opposite sign with probability $p_0$. If $p_0=0$, the first two components are equal for both classes. No information about the class to which the supplied data belong can be thus extracted by comparing the two above components. The second component is indeed a mere copy of the first and therefore irrelevant for classification purposes. On the other hand, if $p_0=1$, the first and second components from class $1$ data are in perfect anti-correlation, while data from class $0$ are still matched and thus fully correlated. 
Hence, in the latter case, it is possible to completely separate the two classes by solely comparing $x_0$ and $x_1$. The Parameter $p_n$ quantifies therefore the degree of sensible information which can be pulled out from a direct comparison of the two components in the $n$-th pair.  We want to stress that, regardless of the value of $p_n$, neither component individually contains information on the specific class to which the vector belongs. Therefore, useful information can be obtained to solve the classification task only by analyzing the two components simultaneously. Following this observations, we generated the dataset by using four different values of $p_n \in [0,1]$ for the four pairs. Finally, the last two components are not correlated with the others and are identically distributed for both classes, hence void of information. We therefore assign a parameter $p_4=0$ to these two components to explicitly state their manifest irrelevance. Note that the two classes are not linearly separable, even in the case one couple of components is associated to a $p_n=1$. A total of $10^5$ examples were generated to perform the test.



 The architecture of the employed neural network, parameterized via its spectral modality, is identical to that put forward for the analysis of the uncorrelated dataset (except for the first layer that in this case receives just 10 entries). In upper right panel of Fig. \ref{fig:corr_multiplot_resuts}, the post-training eigenvalues referred to the input layer are shown. Firstly, we notice that pairs are correctly spotted out, as they are assigned the very same eigenvalue entry. Moreover,  eigenvalues associated with irrelevant components are tuned zero for this dataset as well, an observation that testifies again on the ability of the spectral network to discriminate between irrelevant and relevant features. Lastly, the magnitude of the recorded eigenvalues scales proportionally to the control parameter $p_n$, which quantifies the relevance of the selected pair for classification purposes (bottom right panel, Fig. \ref{fig:corr_multiplot_resuts}). The Pearson correlation ($\rho$) between post-training eigenvalues and the relative $p_n$ values was measured over ten independent training runs, yielding $\rho= 0.988 \pm 0.004$. In conclusion, and also for this second, more complex dataset, the optimized eigenvalues referred to the first layer are a good proxy of the importance of the associated components.

\subsection{The MNIST dataset} 
\label{sec:mnist}

As an additional test, we applied the spectral features detection method to a reduced version of the well-known MNIST dataset \cite{deng2012mnist}. This latter consists of $28 \times 28$ images depicting handwritten digits, ranging from $0$ to $9$, to be eventually classified. In our experiments, and to deliver a clear message, we first focused on the subset of images 
``zeroes" and ``ones". Despite the simplicity of the adopted framework, it is interesting to test the method against a real dataset, where the measure of the relevance of the components is not a priori given. The input vectors of the MNIST dataset has a dimension of $784$, and each component of the vectors represents the intensity of a specific pixel in the corresponding image.

In this context, asking which component is relevant amounts to asking which pixel (or collection of pixels) discloses the information that is used by the trained network to make decision.

\begin{figure}
    \centering
    \includegraphics[width=0.9\textwidth]{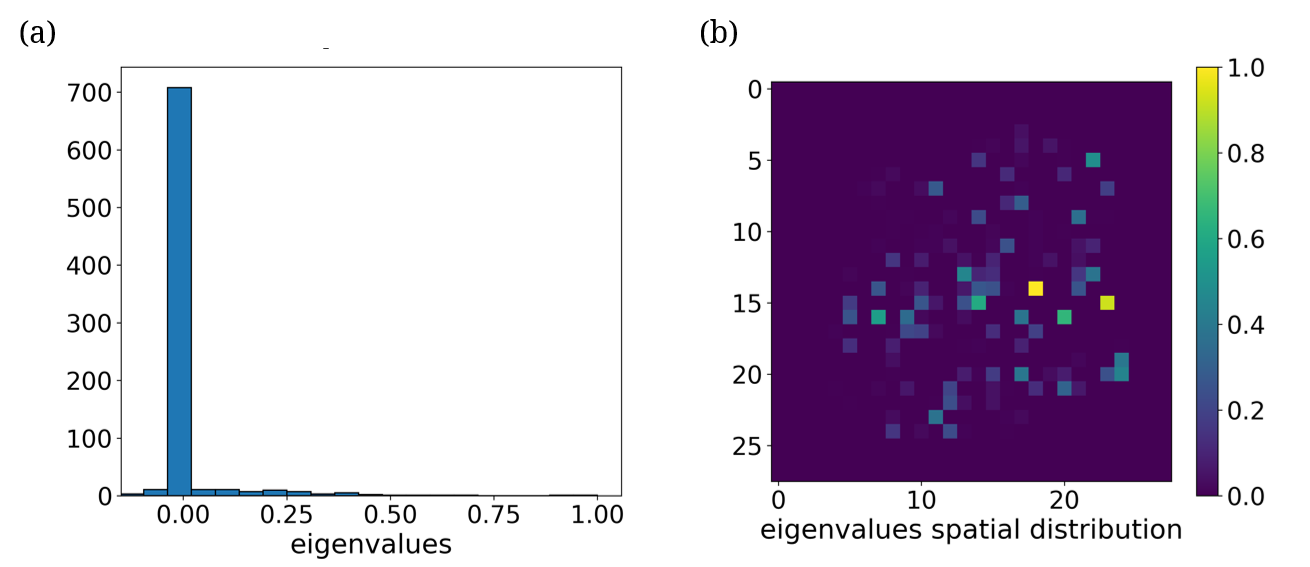}
    \caption{In panel (a) the  distribution of the first $784$ post-training eigenvalues (normalized to their maximum) is reported. Panel (b) shows the input space for the same dataset, where each component (i.e., pixel) is colored according to its relative post-training eigenvalue. }
    \label{fig:mnist_distr}
\end{figure}

Once again, we trained a spectral feedforward neural network by adjusting the trainable spectral parameters. The hidden structure of the network is identical to that employed for the above applications. Clearly, the input layer is now made of 784 elements, the size of the supplied items. The exit layer is still composed of just two nodes, one associated to the images that display a zero and the other pointing to the ones. As expected, the trained model performs very well in terms of estimated classification accuracy (larger than $99$ \%). 

The distribution of the post training eigenvalues associated to the input  layer is reported in panel (a) of Fig. \ref{fig:mnist_distr}. A clear peak at zero appears, signifying that many eigenvalues have been suppressed, thus targeted as ineffective, during training stages. Just a few eigenvalues are non-zero. This conclusion can be rephrased by stating that many pixels are unnecessary for task resolution. Panel (b) of the image \ref{fig:mnist_distr} shows the $28 \times 28$ pixels, the canvas on which images are painted, colored according to the associated eigenvalue score: blue pixels refer to smaller eigenvalues, while yellow pixels stand for large positive values. 

\begin{figure}
    \centering
    \includegraphics[width=0.5\textwidth]{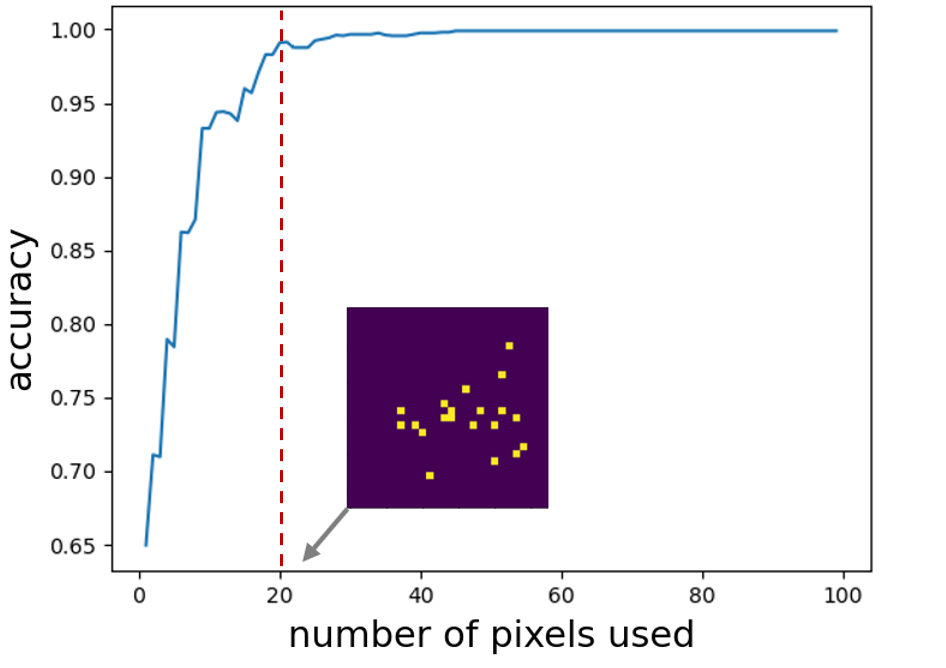}
    \caption{The blue line refers to the accuracy of the model as the number of pixels taken as input grows. The order in which the pixels are inserted follows the absolute value (magnitude) of their corresponding post-training eigenvalues. In the inset, the input space is represented, with the twenty most important pixels highlighted.}
    \label{fig:mnist_accuracy}
\end{figure}

The accuracy of the trained network gets modulated as a function of the number of input pixels included in the analysis, see Fig. \ref{fig:mnist_accuracy}. In this latter Figure, we progressively include more and more pixels ranked from the most to the least important, as according to the spectral (i.e. eigenvalues magnitude) metric. Remarkably
the computed accuracy raises rapidly, with a sharp, almost abrupt approach to asymptotic convergence, a trend that cannot be reproduced by employing any other set  of identical cardinality of randomly selected pixels. 
In particular, by arbitrarily fixing the number of pixels to be included (in descending order of importance) to $20$, one obtains a final accuracy which is competitively close to that recorded when the full load of pixels is provided at the input layer. These 20 pixels, ranked as most important and sufficient for the correct functioning of the trained classification algorithm, can be located back on the 2D grid that defines the images support (inset in Fig. \ref{fig:mnist_accuracy}).

\begin{figure}
    \centering
    \vspace{1cm}
    \includegraphics[width=0.4\textwidth]{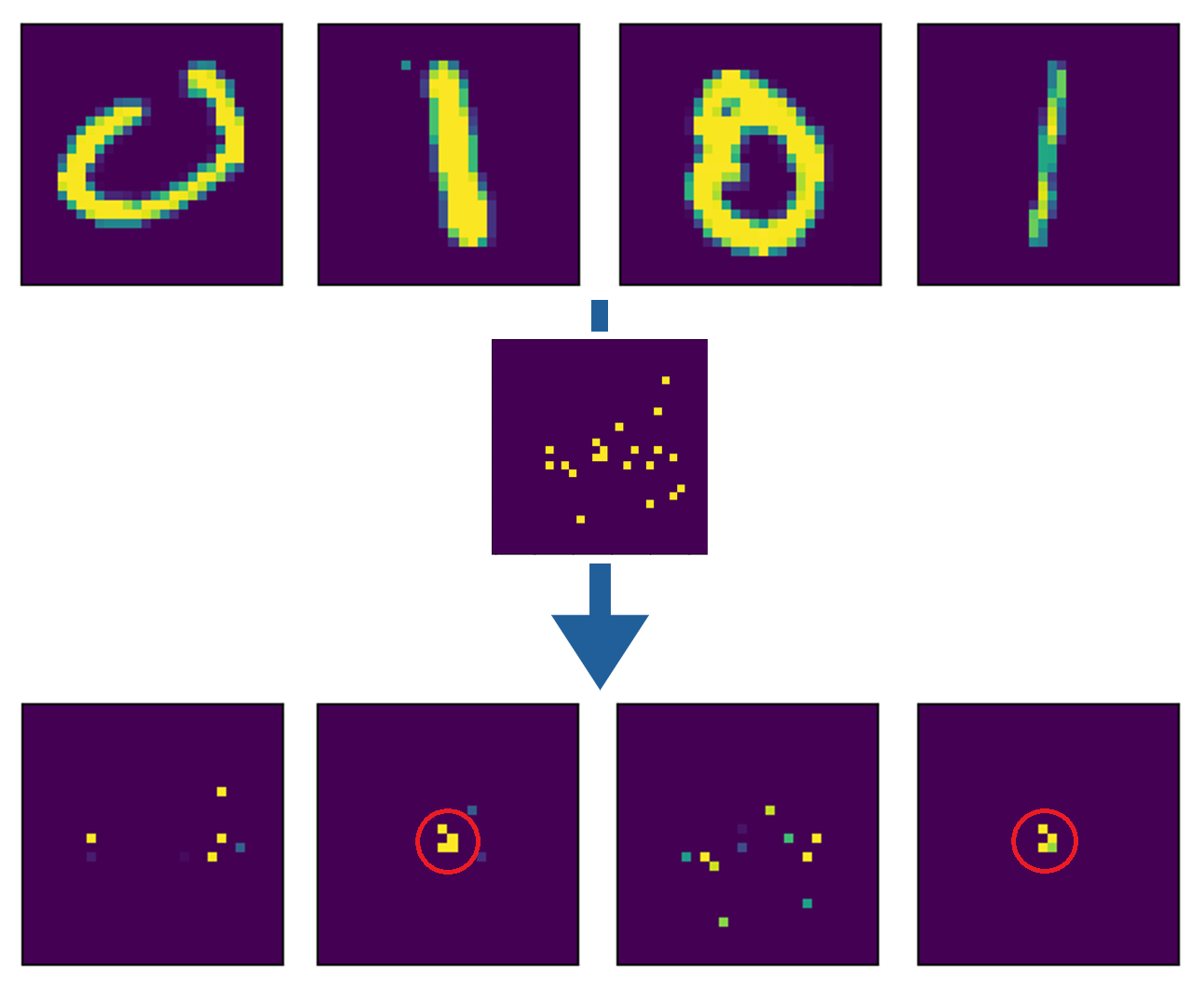}
    \caption{Four examples of images from the MNIST dataset as originally presented (top) and after the application of the mask identified by the proposed method (bottom). The red circles identify the central cluster activated by images representing a  ``one".}
    \label{fig:mnist_mask}
\end{figure}

To figure out how the network exploits the subset of selected pixels for handling the classification task, we can overlay the mask containing only the relevant pixels (the binarized picture displayed as inset in Fig. \ref{fig:mnist_accuracy}) on individual input (images) data. A few examples of what one gets following this procedure are reported in Fig. (\ref{fig:mnist_mask}). Images portraying digit ``one" activate a cluster of pixels positioned in the center of the drawing (red circles), as opposed to images with a ``zero" displayed, which resonate instead with a complementary set of pixels surrounding the center.  The method is hence capable of identifying a minimal subset of input components which prove crucial for the correct handling of the examined data.

\begin{figure}
    \centering
    \vspace{1cm}
    \includegraphics[width=0.7 \textwidth]{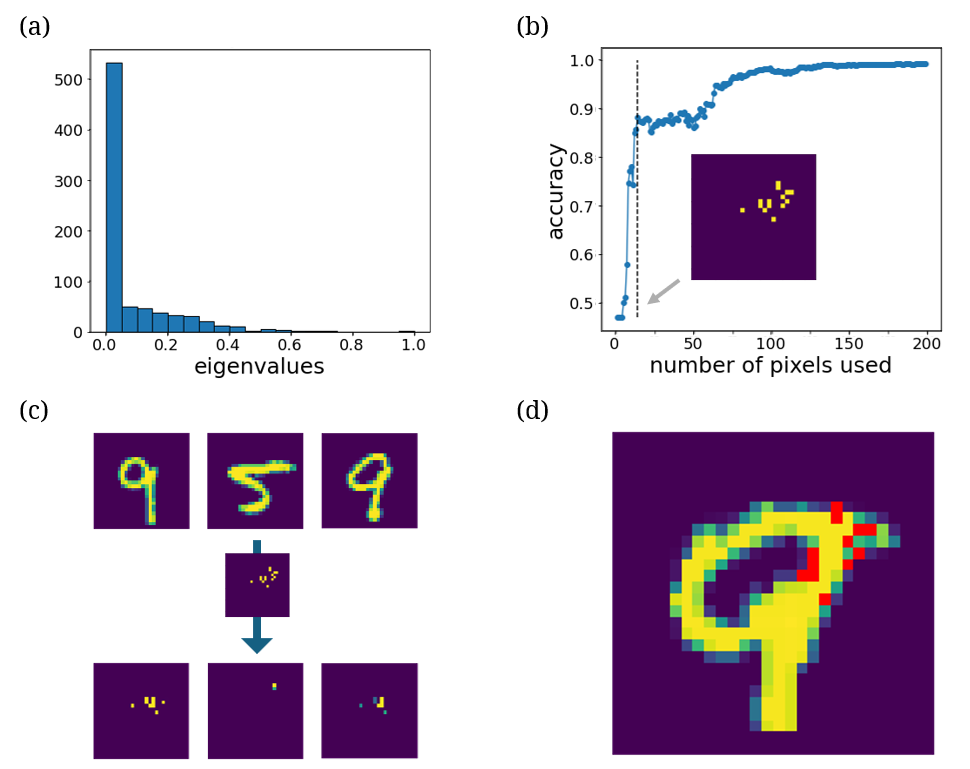}
    \caption{Results for the "five" vs. "nine" case. The neural network was trained to distinguish between images representing the number five and those representing the number nine. Panel (a) displays the distribution of eigenvalues at the end of the training process. A significant peak at zero is observed, indicating that many pixels are not employed for solving the task. Panel (b) illustrates the accuracy trend as a function of the number of pixels provided to the network as input, as follows the reported spectral ranking.  Panel (c) presents three example inputs along with the pixels activated after applying the minimal mask. Panel (d) highlights a specific input case where the pixels belonging to the minimal mask and activated for the considered input are marked in red.}
    \label{fig:mnist_multiplot_5Vs9}
\end{figure}

To further elaborate on the performance of the proposed method for input feature detection, we repeat the above analysis for the subset of images relative to numbers ``five" and and ``nine". The results are in line with those reported above, as it can be readily appreciated by visual inspection of Figure \ref{fig:mnist_multiplot_5Vs9}. With just 14 pixels ($\simeq1.8\% $ of the total) the network  achieves approximately $90\%$ classification accuracy. These pixels are localized in the upper right portion of the image where the "nines" feature a closed circular loop, which is instead lacking in the "fives". This is hence the macroscopic distinctive attribute that is learned by the neural network to deal with the assigned task.

\begin{figure}
    \centering
    \vspace{1cm}
    \includegraphics[width=0.9\textwidth]{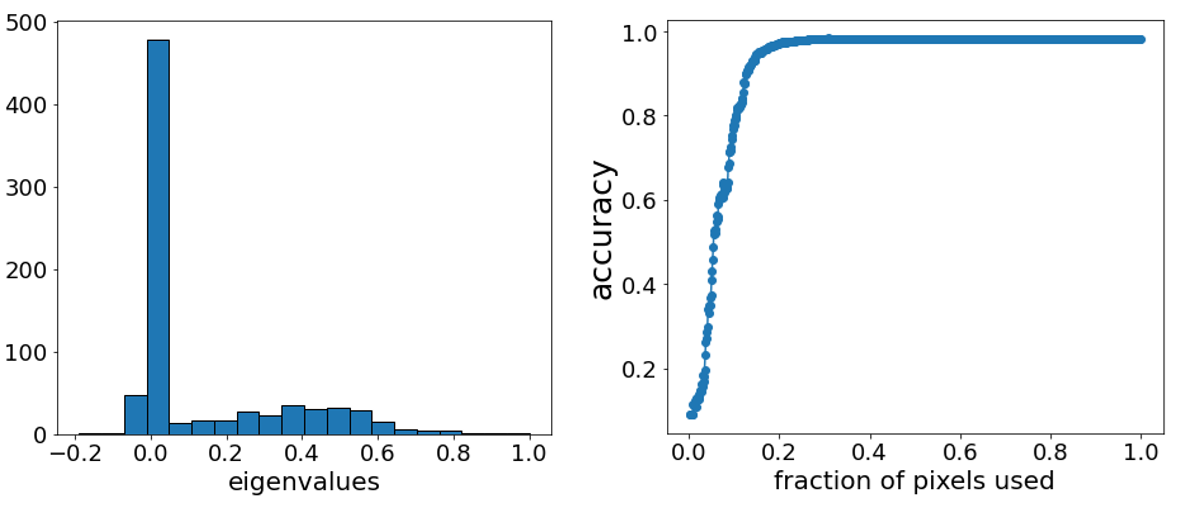}
    \caption{Eigenvalue distribution after training for the full MNIST case (left) and  accuracy of the model as the number of pixels taken as input grows (right). The order in which the pixels are inserted follows the absolute value (magnitude) of their corresponding post-training eigenvalues}
    \label{fig:allmnist_multiplot}
\end{figure}

Finally, we also report in Figure \ref{fig:allmnist_multiplot} the recorded accuracy for the complete collection of digits (ranging from $0$ to $9$) which compose the entire MNIST dataset. Also in this case, and in analogy with the results obtained for the examined settings in which pair of digits are directly compared, the accuracy raises as a function of the number of input pixels included in the analysis, and ranked according to their spectral importance.

\subsection{Stellar spectra and metallicity prediction}
\label{subsec:stellar_res}

\begin{figure}
    \centering
    \vspace{1cm}
    \includegraphics[width=0.9\textwidth]{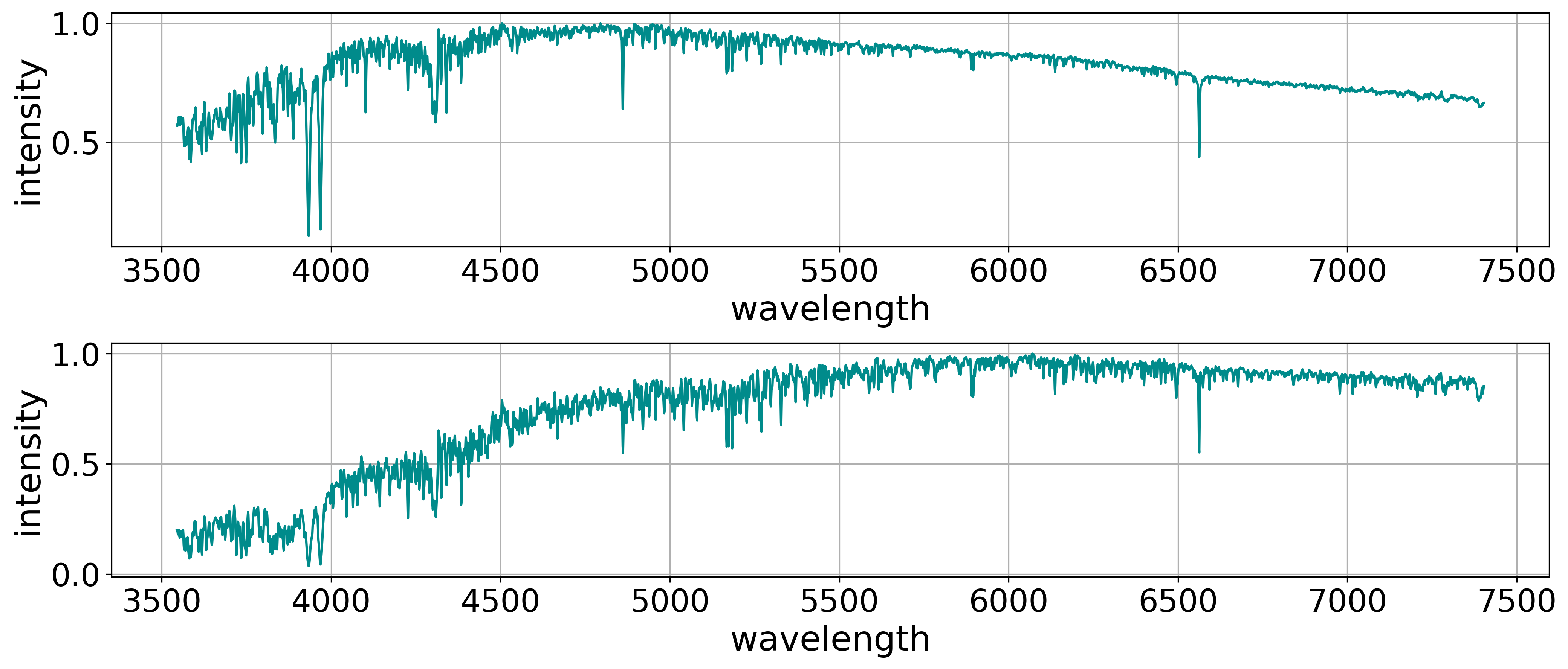}
    \caption{Plot of some of the stellar spectra composing the MILES dataset \cite{sanchez2006medium}.}
    \label{fig:some_stellar_spectra}
\end{figure}

In this Section we report on the performance of the algorithm against a dataset of interest for astrophysics. More specifically, we consider the MILES stellar library \cite{sanchez2006medium}, composed of 985 stars.
The dataset is composed of stellar spectra, namely flux as a function of wavelength, in the range 3525-7500 \AA{}. For each spectrum we consider the associated stellar parameters (effective temperature, gravity, and the abundance of heavy elements known as metallicity) from \cite{cenarro2007medium}. A few representative spectra are shown in Figure \ref{fig:some_stellar_spectra}. Because of its spectral type coverage and flux calibration accuracy this stellar library has been extensively used as a basis for constructing stellar population models to analyze galaxy spectra \cite{vazdekis2010evolutionary, vazdekis2016uv} and therefore deriving metallicity and star-formation histories of galaxies.


To estimate the metallicity, we here train a spectral feedforward neural network, which takes as an input the displayed intensity of the recorded spectra at various wavelengths. The network architecture consists of three hidden layers comprising 200, 50, and 50 neurons, respectively. As in the spirit of the above, the spectral ranking of the supplied input can guide through the identification of a subset of features deemed relevant for an adequate assessment of the star metallicity.

\begin{figure}
    \centering
    \vspace{1cm}
    \includegraphics[width=0.9\textwidth]{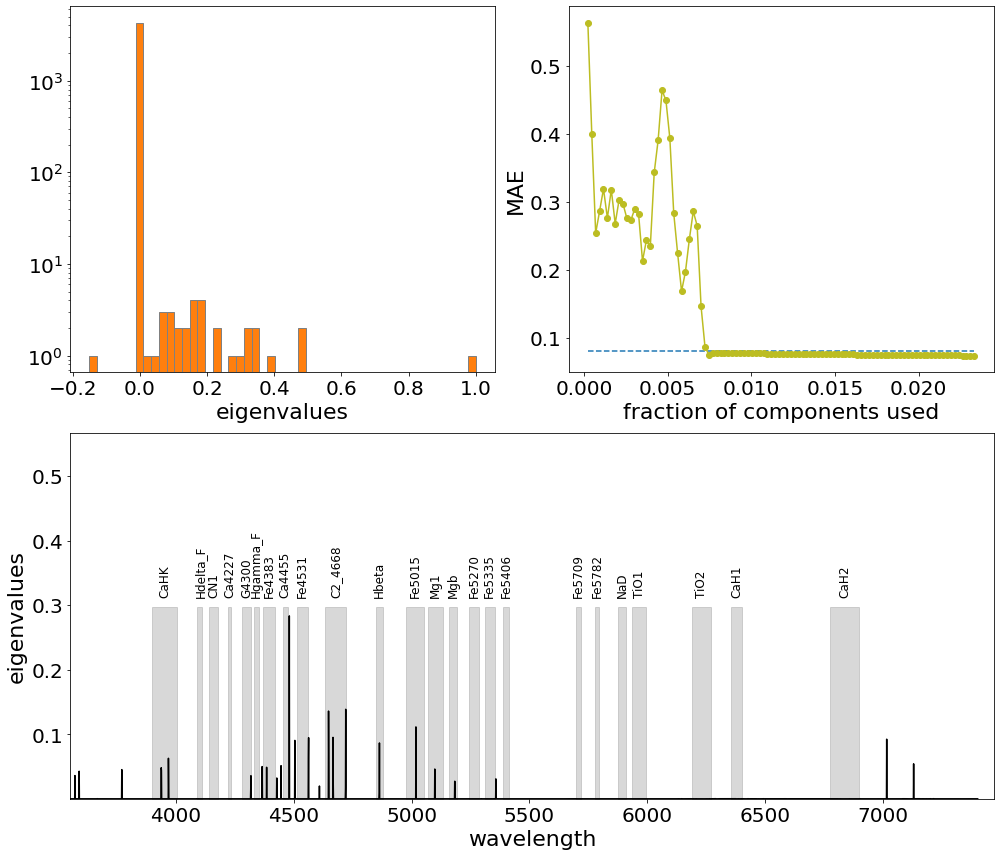}
    \caption{In the upper-left panel  the normalized distribution of the eigenvalues of the input layer for the metallicity prediction task is reported in vertical log scale. Upper-right panel shows the mean absolute error of the model versus the fraction of spectral lines provided as an input. The order in which the lines are inserted follows the absolute value (magnitude) of their corresponding post-training eigenvalues. The dashed line refers to the absolute error made by a standard MLP with the same architecture. The bottom panel shows the wavelength ranges in the stellar spectra which are deemed relevant for the metallicity prediction task (grey shadows) along with the first layer eigenvalues (see also Method section).}
\label{fig:multiplot_stellar}
\end{figure}

We chose metallicity as the target because it can be inferred from well-defined absorption lines of metal species. Such lines are, however, generally blended at the resolution of our dataset and in galaxy spectra. A set of “indices”, or wavelength regions, have therefore been defined \cite{faber1985old,worthey1994old}, which are sensitive to various combinations of metallicity, temperature, and gravity. We will compare the feature importances obtained from the machine learning analysis with these classical indices.

As it can be appreciated by looking at the bottom panel in Figure (\ref{fig:multiplot_stellar}), the most pronounced peaks of $\lambda_i$ (vertical black traits) are aligned to a subset of the defined indices in the stellar spectra (grey shadows). Namely, we observe high importances being associated with the Calcium H and K line (CaII 3934, 3968), Mgb (mostly associated with the MgI triplet 5167, 5173, 5184) and several absorption bands dominated by Fe II (e.g. Fe4384, Fe5015 ). All of these indices are known to be sensitive to metallicity.
From an astrophysical perspective, the Fe bands are generally considered better probes of metallicity than both the Ca H \& K absorptions, which are subject to a number of limitations. Namely, the Ca H \& K lines are known to also vary as a function of age (which affects the H/K ratio), and the abundance of Calcium, which has a different nucleosynthetic origin than Fe. On the other hand, a combination of Mgb and Fe5270 (together with Fe5335) is generally considered the best tracer of total metallicity \cite{thomas2003stellar, candebat2024inferring}, highlighting the success of our network in identifying relevant features. The network attributes some importance to the H$\beta$ absorption line but not to H$\alpha$, as expected given that H$\beta$ has a larger metallicity dependence than the other Balmer lines. However, it is important to note that some spectral lines not typically considered crucial by the astrophysics community for the metallicity prediction task are highlighted as relevant, a result also confirmed by other methods (see Sec. \ref{sec:comparison}).

The upper-left panel in Figure   (\ref{fig:multiplot_stellar}) shows the normalized eigenvalues distribution of the input layer eigenvalues. As for the other examples discussed above, a clear peak is displayed close to zero (note the logarithmic scale on the $y$-axis) thus implying that several input features play no role for the handled task. In particular, if we compute the prediction error on the test set as a function of the number of spectral lines used as input — ordered according to their corresponding eigenvalues — we observe that the minimum error is reached using only around $30$ lines (see the upper-right panel in Figure \ref{fig:multiplot_stellar}). In other words, the network is able to make accurate predictions by relying on just $0.7\%$ of the input components. Notably, the measured error is comparable to that of a standard MLP (see top right panel in figure 11). The eigenvalues allow hence to  identify a subset of relevant input elements that drive decision making across the network. In this respect, the algorithm is thus successfully tested against a real dataset with non trivial physical implications.

\section{Comparison with other methods}
\label{sec:comparison}
In this section, we briefly compare the results obtained using our proposed method with those obtained by employing other approaches, as available in the literature, to carry out the same task. This comparison is not intended to be exhaustive, and a more in-depth analysis between the spectral method in relation to other techniques is left for future work.

\begin{figure}
    \centering
    \vspace{1cm}
    \includegraphics[width=0.9\textwidth]{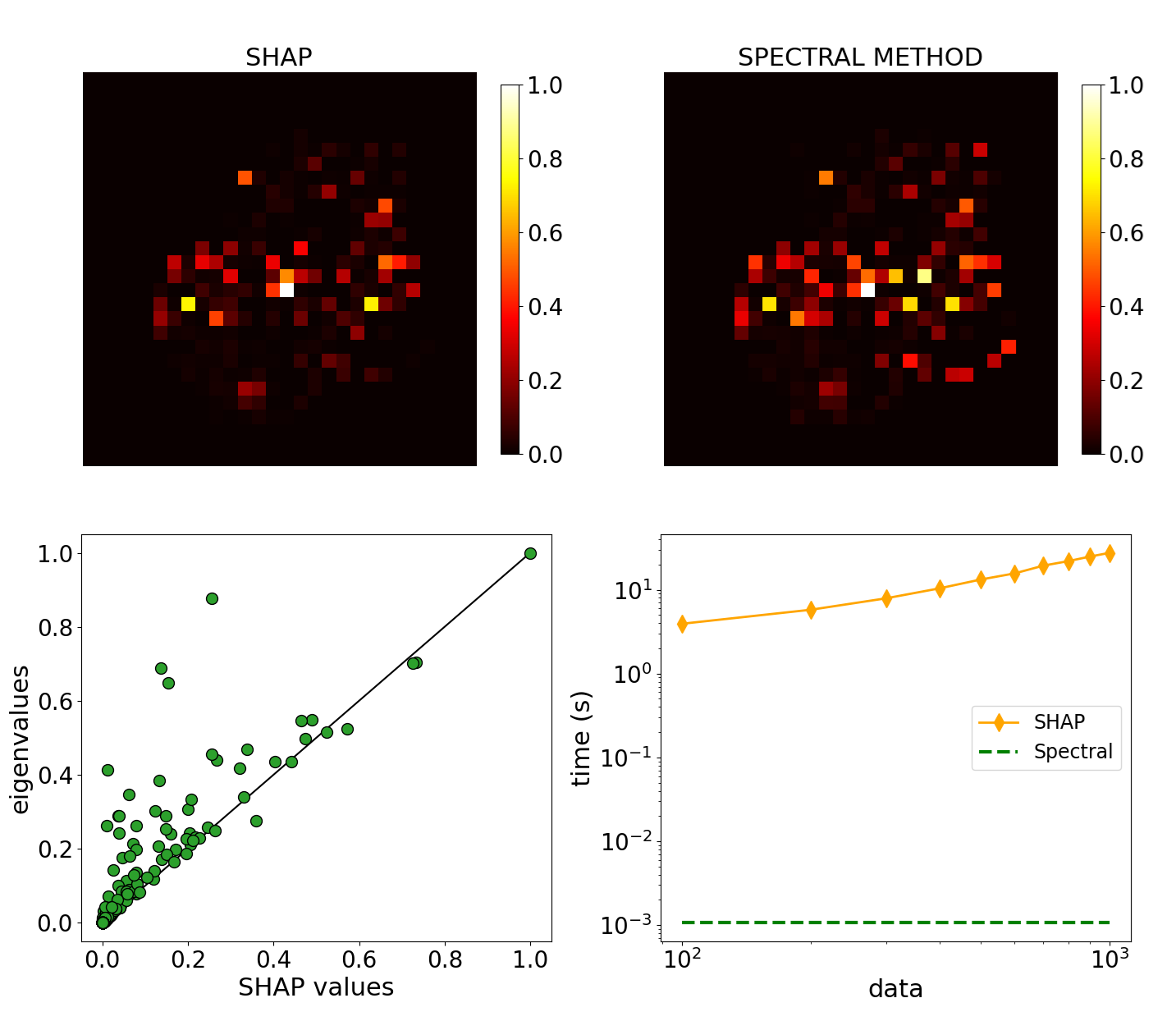}
    \caption{The two upper panels show the importance values obtained by using (i) the SHAP method normalized by their maximum (left) and (ii) the normalized eigenvalues (right). The results refer to a model trained using the spectral procedure on the simplified MNIST dataset (see Section \ref{sec:mnist}), with values represented in the input space. The bottom-left panel plots the normalized eigenvalues against the importance values obtained using the SHAP method. 
    In the bottom-right panel, the computation times required to calculate the normalized eigenvalues (green dashed line) and SHAP values (orange diamonds) are shown for different sizes of the dataset used to estimate the average SHAP scores.}
\label{fig:SHAP_multiplot}
\end{figure}

\textbf{SHapley Additive exPlanations (SHAP).}
A first comparison was made with the SHAP method \cite{lundberg2017unified}, a well-known framework based on game theory which is able to explain how input features contribute to the model's prediction. Since our method provides a measure of the global relevance of input components, independent of any specific item being inspected, we derived comparable information by averaging the importance values obtained by using SHAP over multiple data points.
First of all, we trained a model using the spectral procedure on the simplified MNIST dataset, as described in section \ref{sec:mnist}. We then compared the normalized eigenvalues (i.e., the eigenvalues multiplied by the norm of the corresponding eigenvectors) as obtained at the end of the training with the average pixel importance values provided by SHAP on the same model. In the upper panels of Figure (\ref{fig:SHAP_multiplot}), the relevance values estimated by using the two methods are shown in the input space. The results are very similar: the pixels identified as most relevant are distributed in the same way within the image. In the bottom-left panel, the normalized eigenvalues are plotted against the SHAP-based relevance scores, showing good agreement between the two estimators. This confirms that post-training eigenvalues provide reliable estimates of the input component importance, comparable to the relevance scores produced by the SHAP approach. However, since eigenvalues are parameters tuned during the training phase and are inherently input-independent, the computational cost of the spectral method is significantly lower. To prove this, in the bottom-right panel of Figure (\ref{fig:SHAP_multiplot}), the computation time required to obtain SHAP relevance scores is reported for varying sizes of the portion of the dataset used for averaging. As the subset size increases, the computation time scales linearly. Nevertheless, even for very small subsets, the SHAP computation time is orders of magnitude higher than the time required to normalize the eigenvalues of the first layer, which is the only post training operation required when dealing with the spectral method.

\begin{figure}
    \centering
    \vspace{1cm}
    \includegraphics[width=0.9\textwidth]{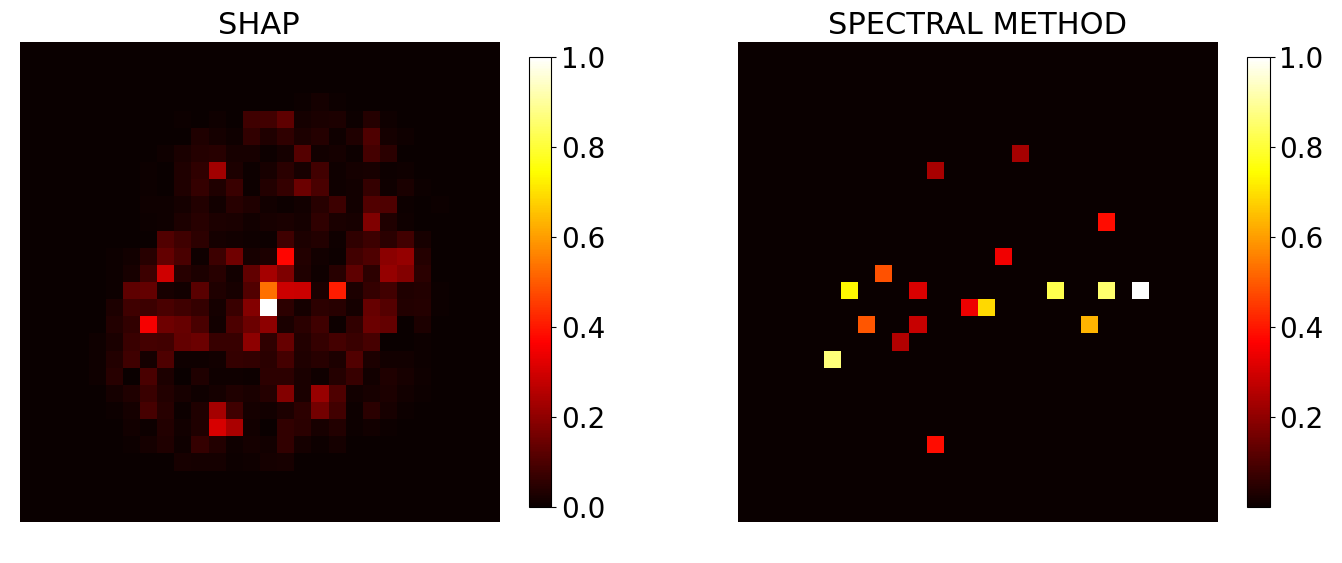}
    \caption{The left panel shows the importance values obtained using the SHAP method normalized by their maximum (left) and for a standard MLP with a $L_1$ regularization factor on the weights of the first layer. The right panel shows the normalized eigenvalues of the first layer of a spectral neural network where a $L_1$ regularization factor has been used on the eigenvalues. Both the results refer to the simplified MNIST and the values are plotted in the input domain.}
\label{fig:SHAP_direct_l1}
\end{figure}

It is important to note that methods such as SHAP are limited to providing an estimate of the importance of input features, but do not include strategies for guiding models toward solutions that rely on few input components. Such strategies must be formulated separately. In fact, when examining the scores obtained using the SHAP approach on a standard (non-spectral) model with the same architecture and hyperparameters, it can be observed that the scores are nonzero across most of the input domain, even when a regularization term is applied to the weights of the first layer (Figure \ref{fig:SHAP_direct_l1}, left).
In contrast, using the spectral formulation, it is possible to guide the model during training toward solutions that are as sparse as possible in terms of the input components used. This is achieved by introducing a regularization term on the eigenvalues of the input layer. To emphasize this, we apply an $L_1$ regularization on the eigenvalues, enforcing solutions that are even sparser than those discussed earlier in this section and in Section (\ref{sec:mnist}) (Figure \ref{fig:SHAP_direct_l1}, right).



\textbf{Layer-Wise relevance propagation}. The second technique against which we compare our proposed method is Layer-wise Relevance Propagation (LRP) \cite{montavon2019layer}. More specifically, we do not present here a comparison of the respective experimental results. Instead, we elaborate on the theoretical framework underlying the LRP method and explain how it can be naturally reconciled with the spectral parameterization. Let us begin by recalling the basic aspects that underlie the LRP method. The LRP technique aims to estimate the relevance $R^l_i$ of nodes by propagating relevance scores backward from the output to the input. The relevance $R^l_i$ of a node in layer $l$ is related to the relevance $R^{l+1}_j$ of the nodes in layer $l+1$ through the linear relationship:
\begin{equation}
    R^l_i = \sum_j \frac{z_{ij}}{\sum_k z_{kj}} R^{l+1}_j.
    \label{eq:RP}
\end{equation}
The elements $z_{ij}$ define how the relevance scores are distributed when they are back-propagated from a layer to the previous one. Different choices for such elements and for the employed normalization term appearing in the denominator yield different rules for LRP. Some common rules are:
\begin{equation*}
    \text{LRP-0 rule:}\hspace{20mm} R^l_i = \sum_j \frac{a_i w_{ij}}{\sum_k a_k w_{kj}} R^{l+1}_j,
\end{equation*}
\begin{equation*}
    \text{LRP-$\epsilon$ rule:}\hspace{20mm} R^l_i = \sum_j \frac{a_i w_{ij}}{\epsilon +\sum_k  a_k w_{kj}} R^{l+1}_j,
\end{equation*}
\begin{equation*}
    \text{ LRP-$\gamma$ rule:}\hspace{20mm} R^l_i = \sum_j \frac{a_i (w_{ij}+ \gamma w_{ij}^+)}{\sum_k a_k (w_{kj}+ \gamma w_{kj}^+)} R^{l+1}_j,
\end{equation*}
where $w_{ij}$ is the weight of the connection between neurons $i$ and $j$, and $a_i$ is the activation observed at neuron $i$ when the input data is provided to the model.  This clearly implies that the method is input-dependent. Furthermore, different LRP rules can be derived assuming a first order Taylor expansion of the Deep Neural Network around an Input Dependent point, therefore specifically intertwining the value of the back-propagated relevance and the input.

These rules can be rewritten under the assumption of a spectral parameterization by substituting $w_{ij} \rightarrow  \lambda_i \psi_{ij}$. Focusing on the LRP-0 rule, we have:
\begin{equation}
    R^l_i = \sum_j \frac{a_i \lambda_i \psi_{ij}}{\sum_k a_i \lambda_k \psi_{kj}} R^{l+1}_j =\lambda_i  \sum_j \frac{a_i \psi_{ij}}{\sum_k a_i \lambda_k \psi_{kj}} R^{l+1}_j,
\end{equation}

from which it is clear that the relevance $R^l_i$ of node $i$ is modulated by the corresponding eigenvalue $\lambda_i$. This result can be easily extended to $\epsilon$-rule. Moreover, by assuming $\lambda_i > 0$ $\forall i$, the result holds also for the $\gamma$-rule, in fact:
\begin{equation}
    a_i (w_{ij}+ \gamma w_{ij}^+) = a_i (\lambda_i \psi_{ij}+ \gamma \lambda_i \psi_{ij}^+) \propto  \lambda_i.
\end{equation}
The condition $\lambda_i>0$ is typically satisfied since the eigenvalues are initialized to one (see Methods), but it can also be enforced as a constraint during training.
Finally, the global relevance of a node (i.e., input-independent) provided by the LRP method can be computed by averaging over multiple data points. However, the eigenvalues are not affected by this averaging process, and the global relevance of a node remains modulated by its corresponding eigenvalue, namely:
\begin{equation}
    \langle   R^l_i \rangle \propto \lambda_i,
\end{equation}
where $\langle   R^l_i \rangle $ denotes the average relevance of node $i$ across the dataset.
A complete analysis of the consequences of this relation is left for future work. 
Once again, as for the SHAP method, it is important to emphasize that LRP is an input-dependent method and requires extensive post-training computations to estimate relevance. Moreover, it does not provide any mechanism for guiding the model toward solutions that rely on a minimal set of input components, as it is instead possible via the spectral paradigm complemented with suitable regularization terms.

\textbf{Gradient-based methods.} Finally, we compare our method with some gradient-based techniques. In particular, we considered two distinct approaches: the first is essentially a vanilla saliency map, where the importance of the input features is directly associated with the gradient of the output with respect to the components of the input vector. The second is a more advanced technique known as Integrated Gradients \cite{sundararajan2017axiomatic}, which we here choose to apply by setting a zero vector as the baseline. Since both methods provide input-dependent relevance estimates, we averaged the results over multiple elements to obtain a global relevance estimate.
\begin{figure}
    \centering
    \vspace{1cm}
    \includegraphics[width=0.9\textwidth]{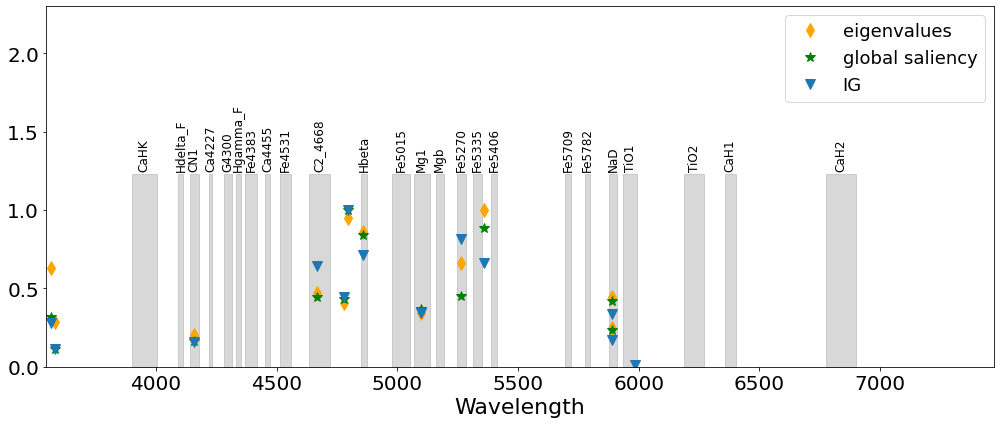}
    \caption{Wavelength ranges in the stellar spectra which are deemed relevant for the metallicity prediction task (grey shadows) along with the first layer eigenvalues (orange diamonds), saliency values (green stars) and importance values given by integrate gradients (blue triangles). Only the values larger than $10^{-2}$ are reported.}
\label{fig:IG_spectra}
\end{figure}
In Figure (\ref{fig:IG_spectra}), we report the results obtained for the stellar spectra dataset, as introduced above. The model was trained using the spectral formulation with $L_1$ regularization applied to the eigenvalues and eigenvectors of the input layer. The three methods (spectral, saliency map, and integrated gradients) identify the same wavelengths as essential for solving the regression task, assigning comparable importance scores to those wavelengths. Indeed, the correlation between the post-training eigenvalues and the saliency map scores for this task (computed over ten training runs of the model) is $0.98 \pm 0.01$, while the correlation with the integrated gradients scores is  $0.95 \pm 0.02$. We have also measured the computation time required to obtain the global score using the three different methods during the same training runs. The spectral method proved to be the fastest, with a computation time of $t = 0.003 \text{s} \pm 0.0005 \text{s}$. Since an average over multiple samples is required to compute the other two global (input-independent) scores, their computation times depend on the number of samples over which the average is performed. Even when averaging over only $20$ samples (the size of the test set for this task), the computation time for the global saliency map scores is $t = 0.34 \text{s} \pm 0.03 \text{s}$, while for the integrated gradients it is $t = 0.58 \text{s} \pm 0.02 \text{s}$. A more exhaustive analysis of the computational time and algorithmic complexity is reported in Appendices (\ref{apdx:method_comaprison_C}) and (\ref{apdx:algo_complexity_D}).). In any cases, the observed reduction in the number of relevant components is a consequence of applying regularization during training to the new parameters introduced by the spectral formulation, namely the eigenvalues, a possibility that is not available when working with a standard MLP.

From all these comparisons, we can conclude that the post-training eigenvalues of the first layer provide a sensible ranking of the input components, in line with that recovered with other available techniques. This demonstrates that the eigenvalues serve as a reliable proxy for the global (i.e., input-independent) relevance of the input components. Moreover, we showed how the use of spectral formulation is an intrinsically different approach compared to the methods described above. In fact, it not only provides a means to estimate relevance, but also offers a strategy to easily enforce solutions that rely on sparse representations of the data. Let us sum up some pros and cons of the proposed method:
\newline

\noindent\textbf{Pros:}
\begin{itemize}
    \item \textit{Input-independent method.} The eigenvalues serve as an effective proxy for the relevance of the input components across the entire dataset. Furthermore, no additional computation is required to aggregate these relevance measures over multiple data instances.
    \item \textit{Very fast algorithm}. The only computation required to obtain the relevance on the input components is to normalize the first-layer eigenvalues (i. e.,  to multiply them by the norm of the associated eigenvectors).
    \item \textit{Enforce sparse solutions}.  As the parameters approximating feature relevance are trainable components of the model, they can be regularized to favour solutions that rely on a minimal set of input components.
\end{itemize}

\noindent\textbf{Cons:}
\begin{itemize}
    \item \textit{Input-independent method.} Having an input-independent method can be disadvantageous when one wants to understand the decision-making process that led to a prediction for a specific input. A possible solution is to overlay the global relevance map onto the specific input to extract data-dependent importance values, as reported in Figure \ref{fig:mnist_mask}.
    \item \textit{Model-dependent.} The method applies solely to neural network models formulated with spectral parameterization.
\end{itemize}

\section{Discussion}

We have here elaborated on a procedure that allows one to easily extract input feature relevance in a neural network model and to force solutions relying on sparse representations. The procedure is input-independent and it is founded on solid mathematical grounds. The assessment on the actual relevance of the provided input feature follows the evaluation of a scalar indicator (one for each of the features) that self-consistently gauges the impact of the supplied information in the decision making problem. These parameters are mathematically traced back to the eigenvalues of the transfer operator linking the first two layers. The eigenvalues are trainable parameters and they are adjusted during the training to solve the task under consideration. During training, it is possible to regularize these parameters to enforce solutions making use of a minimal set of the input components.

Based on the above reported analysis carried out for both mock and real data, we can positively conclude on the adequacy of the proposed features' detection algorithm. The eigenvalues associated to the input nodes of the collection provide indeed a sensible ranking of the input features. The relevance of input components obtained with the proposed method has been compared with that obtained working with different well-known methodologies showing good agreement. 

By working with a relatively simple dataset, we could elaborate on a sound explanation,  drawn a posteriori, of the process that underlying decision making. This includes to {\it visually} isolate the relevant patches of information that are selectively {\it seen} by the trained device in operation mode. For more complex datasets, the posterior explanation could be less transparent in terms of grasping the qualitative message delivered. As an obvious corollary of the analysis, the method could be also applied as a compression algorithm to store large data in a minimal memory kernel.

\section{Methods}

\textbf{Spectral Formulation}
We provide here a few additional details on the spectral approach to the training \cite{giambagli2021machine}. As mentioned in the main body of the paper we consider layer $l-1$ and layer $l$ , respectively made of $N$ and $M$ nodes. The state of the system on layer $l-1$ is encapsulated in vector $\textbf{x}^{l-1}=\left(x^{l-1}_1,\hdots,x^{l-1}_N\right)$. The state on layer $l$ is instead stored in vector $\textbf{x}^{l}=\left(x^{l}_1,\hdots,x^{l}_M\right)$. The mapping is implemented as $\textbf{x}^l = f(W\textbf{x}^{l-1})$, where $W$ is a $M\times N$ matrix and $f$ is a non-linear funtion. The spectral recipe deals with the equivalent transformation $\textbf{z}^{l}= f(A \textbf{z}^{l-1})$, where
$\textbf{z}^{l-1} = \textbf{x}^{l-1}=\left(x^{l-1}_1, x^{l-1}_2, \hdots x^{l-1}_{N}, 0, 0, \hdots 0 \right)$ and the last $M$ elements of vector $\textbf{z}^{l}$ refer to the actual activity displayed on 
layer $l$. An appropriate projection matrix should be formally considered at the end of each mapping between consecutive layers, which effectively implements the transfer of
the activations from bottom to upper elements (while filling the bottom portion of the vector with
zeros) of the next vector to be processed.

As anticipated in the main body of the paper $A=\Phi \Lambda \Phi^{-1}$, where $\Phi$ and $\Lambda$ are $M\times N$ matrices. The first identifies the eigenvectors of $A$, the latter the associated eigenvalues. More explicitly we assume:

\begin{equation}
    \Phi = \begin{pmatrix}
        \begin{matrix}
            \mathbb{I}^{N\times N}
        \end{matrix}  &    
         \begin{matrix}
            0
        \end{matrix}\\
        \begin{matrix}
            \Psi
        \end{matrix} &
        \begin{matrix}
            \mathbb{I}^{M \times M}
        \end{matrix}
    \end{pmatrix},
\end{equation}

and:

\begin{equation}
    \Lambda = \begin{pmatrix}
        \lambda_1 &  & && \\
         & \lambda_2 &  & &  \\
         &  & \ddots & &  \\
         &  &  &  & \lambda_{N+M} \\
    \end{pmatrix}.
\end{equation}

As already recalled it can be proven that $\Phi^{-1} = 2\mathbb{I} -\Phi$  \cite{buffoni2022spectral, chicchi2021training}. By making use of this latter expression, it is immediate to show that:

\setlength{\fboxsep}{1.5em}
\begin{equation}
    A = \begin{pmatrix}
        \begin{matrix}
        \lambda_1 &  &  &  \\
         & \lambda_2 &  &  \\
        & & \ddots&  \\
         & &  &\lambda_N  \\
    \end{matrix} &0\\
     \fbox{W}  &\begin{matrix}
        \lambda_{N+1} &  & & \\
         &  \ddots & \\
         &  &\lambda_{N+M} \\
    \end{matrix} 
    \end{pmatrix}.
\end{equation}

The part of $A$ responsible of the information transfer from 
the departure layer $l-1$ to the destination layer $l$ is the bottom-left block $W$. The relation between the elements of $W$ and the spectral parameters (i.e., the non-zero elements of the matrix $\Lambda$ and the elements of the $\Psi$ block of the eigenvector matrix) can be analytically computed yielding
\begin{equation}
 w_{i\rightarrow j} = w_{ji} = (\lambda_i - \lambda_j)\psi_{ji}.
\label{eq_spectralparam_app}
\end{equation}

The weights $w_{ij}$, (associated in the standard formulation to the links of the feedforward network to be trained) are now functions of the spectral attributes of matrix $A$. Following this alternative parameterization, the target of the optimization are the components and the eigenvectors and the corresponding eigenvalues. From optimized eigenvectors and eigenvalues one can compute the weight of the transfer matrix in direct space via equation (\ref{eq_spectralparam_app}). The markers used for discriminating relevant features are the eigenvalues of $A$ rescaled by a quantity that accounts for the norm of the associated eigenvectors. In formulae, $\tilde{\lambda}_i = \lambda_i \left( \sum_j \phi_{ji}\right)^{1/2}$. All along the paper we drop the tilde and refer to $\tilde{\lambda}_i$ as to the $i$-th eigenvalue of matrix $A$.

\textbf{Regularization}
During training we employ an $L_1$ ora a $L_2$ regularization loss on the spectral parameters, $\phi$ and $\lambda$. This could also be understood as an implicit minimization of the Frobenius norm of the operator $W=\lambda \odot \phi$ as $||W||_F = \sum_{i,j}\lambda_i^2\phi_{ij}^2$, therefore learning a low rank representation of the operator. 

\textbf{Training procedure}
The optimization process for all four datasets discussed in the main text was conducted using the Adam optimizer \cite{2015-kingma}. The learning rates were set to 0.01 for the first two datasets, 0.0001 for MNIST and 0.001 for the stellar dataset. Sparse Categorical Crossentropy was employed as the loss function for the  three classification tasks, while the mean absolute error was used as loss function for the metallicity prediction task. For all tasks under consideration, we applied $L_2$ regularization to both the eigenvalues and eigenvectors of the first layer. We also tested $L_1$ regularization when it is specified in the text. The regularization coefficients were set to $10^{-2}$ for the eigenvalue term and $10^{-4}$ for the eigenvector term. The eigenvalues are initially set to one, following the initialization rule presented in \cite{giambagli2024student}.  The number of epochs was set to 1000 for the Gaussian dataset, 500 for the correlated Gaussian dataset, 5000 for MNIST and 1000 for the stellar spectra dataset. Using a large number of training epochs allows us to significantly reduce the regularization term associated with the eigenvalues in the loss function. This, in turn, enables the model to converge to a solution that relies on a minimal subset of input features while still correctly classifying the data. Such sparsity in the input improves the explainability of the model.
It is important to note that employing a large number of epochs is not a stringent requirement: even when operating with fewer training epochs or smaller regularization terms, the eigenvalues of the input layer remain valid indicators of node relevance. However, under these conditions, the model tends to distribute importance across a larger number of input components. While this still yields accurate classification, the explanation of the process becomes more diffuse and less localized.

\begin{figure}
    \centering
    \vspace{1cm}
    \includegraphics[width=0.9\textwidth]{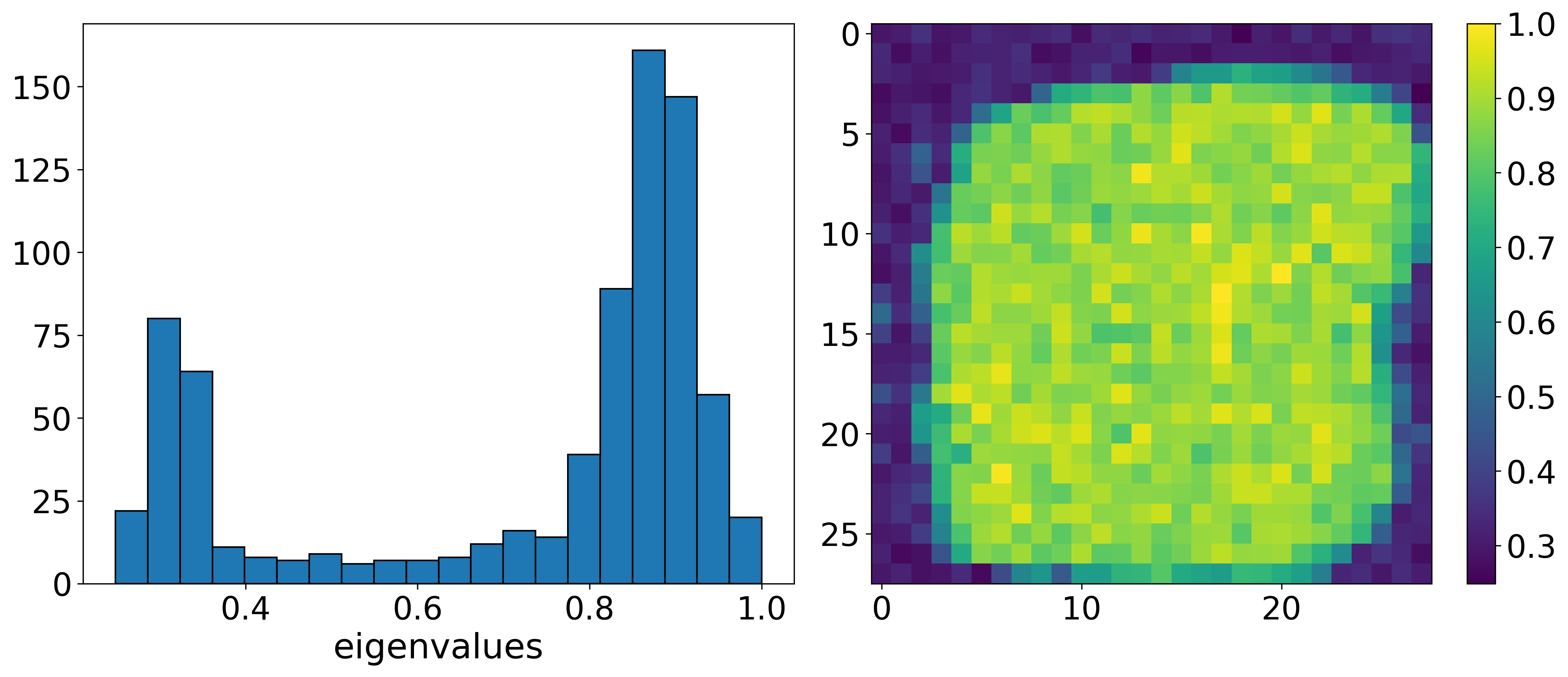}
    \caption{Distribution of the first layer eigenvalues for the MNIST dataset for a model trained with a smaller number of epochs (left) and spatial distribution of the eigenvalues on the input space (right).}
    \label{fig:mnist_less_epochs}
\end{figure}
\begin{figure}
    \centering
    \vspace{1cm}
    \includegraphics[width=0.9\textwidth]{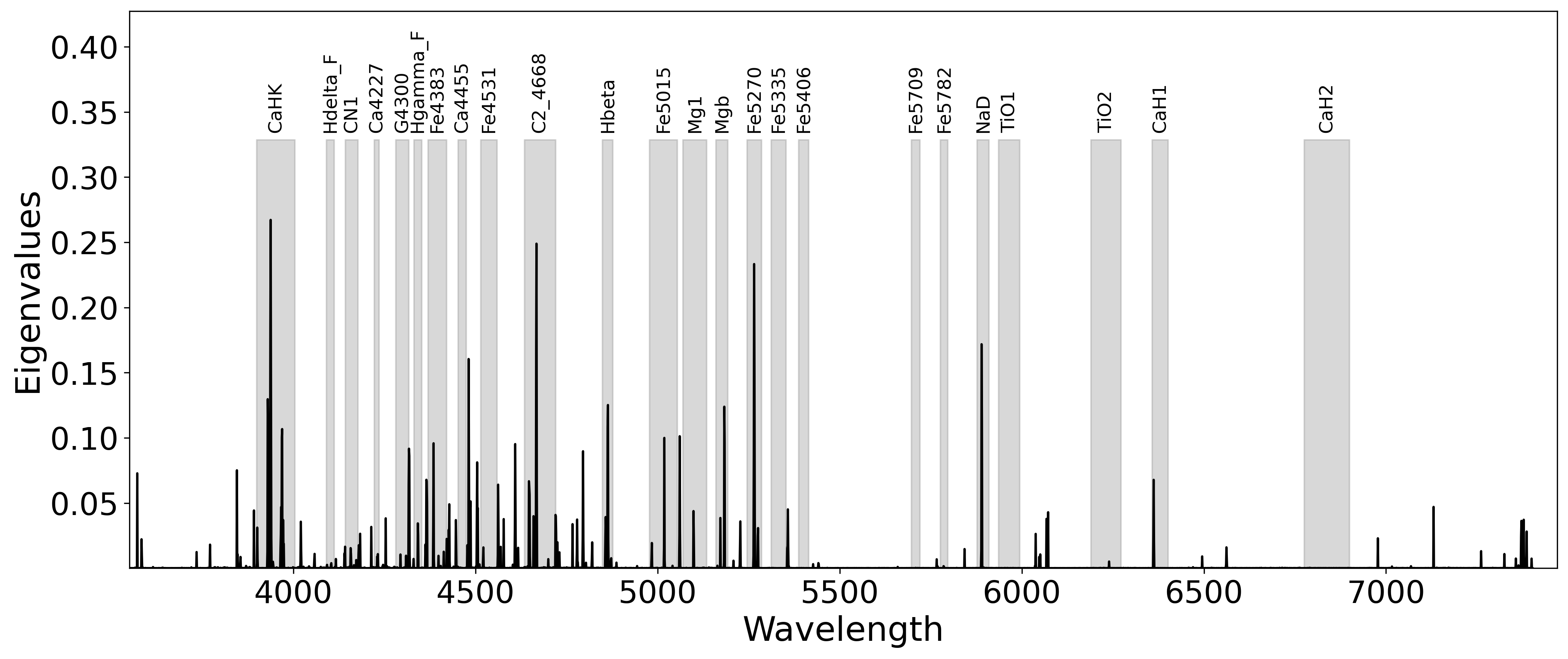}
    \caption{First layer eigenvalues obtained with fewer training epochs and a smaller regularization coefficient (black vertical lines), shown alongside the indices in the stellar spectra considered relevant for the metallicity prediction task (grey shadows)}
    \label{fig:spectra_less_epochs}
\end{figure}

As empirical evidence of the above, we report the results for the MNIST dataset and the stellar dataset as obtained for a smaller number of training epochs and reduced regularization factors. By significantly reducing the number of epochs during training on the MNIST dataset, we find a bimodal distribution for the eigenvalues that refer to the input layer with a peak close to zero that is less pronounced as compared to that reported in Sec. (\ref{sec:mnist}), and a prominent bulk in the region of larger values (See left panel of Fig. \ref{fig:mnist_less_epochs}). This indicates that a greater number of input components are being used to solve the task. By examining the spatial distribution of the eigenvalues, we observe that the relevant components are spread across nearly the entire image (See right panel of Fig. \ref{fig:mnist_less_epochs}). 
In Figure (\ref{fig:spectra_less_epochs}) the results obtained for the stellar dataset are displayed: the eigenvalues associated to the input layer are plotted (black vertical lines) for a model trained with 100 epochs and a regularization coefficient for the eigenvalues  set to $5\times 10^{-4}$. The test loss at the end of the training is essentially the same as that obtained for the case reported in Sec. (\ref{subsec:stellar_res}) (not shown). However, the number of non-zero eigenvalues in the first layer is larger, although most of the peaks still lie in spectral regions known to be crucial for metallicity prediction. In other words, the network detected another equivalent solution that makes use of a larger set of input components.

We also evaluated the effect of different settings for other hyperparameters, such as the number of layers, the use of batch normalization, and the application of dropout. We did not observe any significant changes in the results under these variations. The eigenvalues of the input layer are parameters directly tied to the input components and are unaffected by operations applied in the subsequent layers of the network. For these reasons, our method demonstrates strong robustness to architectural changes and to the use of techniques like batch normalization or dropout. Therefore, the hyperparameters were manually optimized from a restricted set of possible candidates by assessing the model's performance on a validation set ($10$ \% 
of the training set).

\bibliography{bib_new}

\section*{Data availability}
A public repository has been created containing the code used during the work: \url{https://github.com/Lorenzochicchi/SpectralNNs_for_Input_Relevance}. Specifically, the code for generating the synthetic datasets used, the code for training the models, and the code for plotting the images included in the work are provided in the form of notebooks.

\section*{Acknowledgements}

This work is supported by \#NEXTGENERATIONEU (NGEU) and funded by the Ministry of University and Research (MUR), National Recovery and Resilience Plan (NRRP), project MNESYS (PE0000006) "A Multiscale integrated approach to the study of the nervous system in health and disease" (DR. 1553 11.10.2022). We thank Francesco Belfiore and Michele Ginolfi for discussion about the astrophysical implications of our findings.

\appendix

\section{Gaussian dataset with non-isotropic covariance}
\label{appndx:A}
\begin{figure}
    \centering
    \vspace{1cm}
    \includegraphics[width=0.9\textwidth]{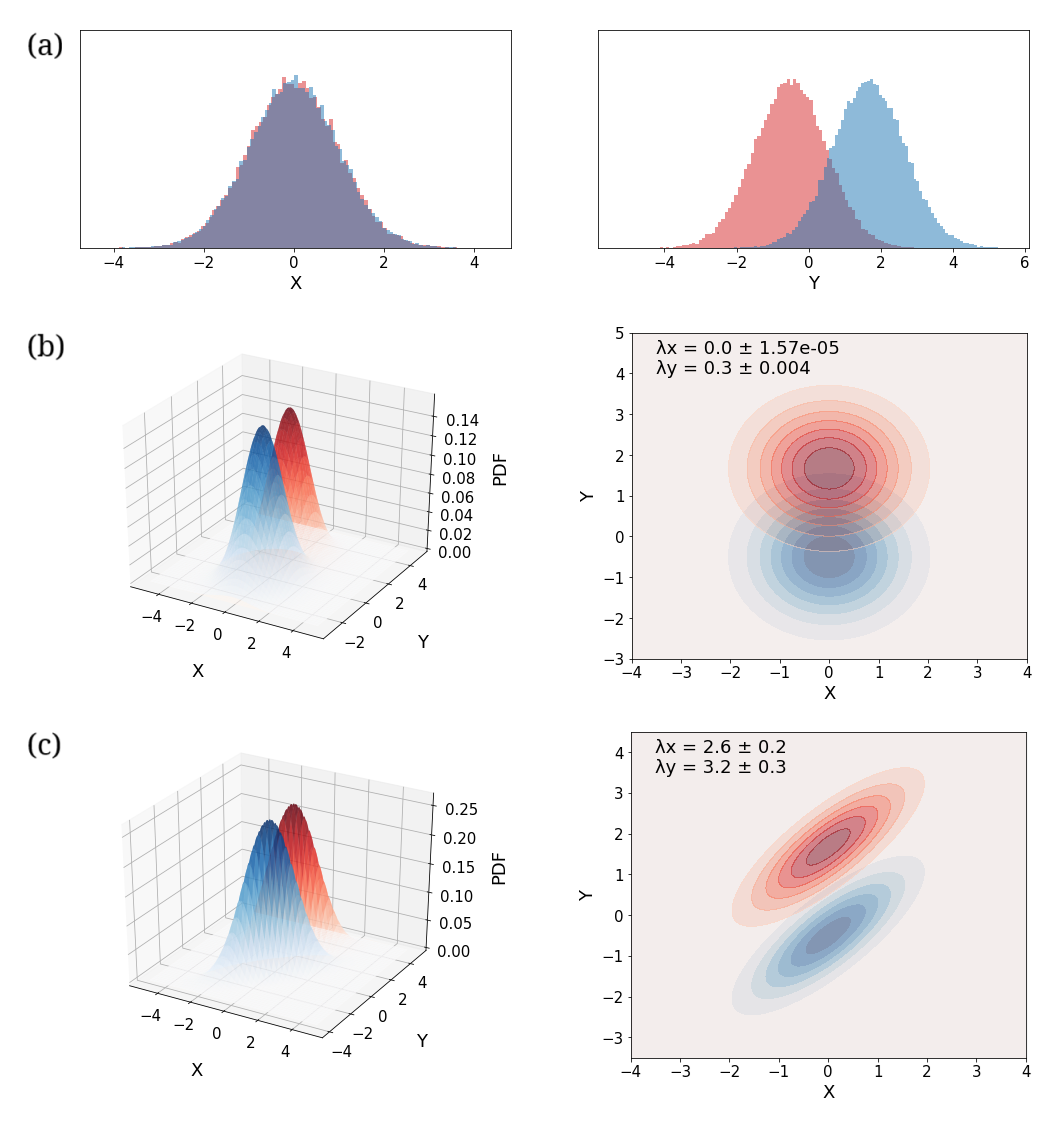}
    \caption{In panel (a), the marginal distributions for components $x$ and $y$ are shown. This situation could, in principle, correspond either to the case of uncorrelated components (panel (b)) or to that of correlated components (panel (c)). In the first scenario, component $x$ is irrelevant for solving the task, and its corresponding eigenvalue is set to zero after training (see the inset in the right image of panel (b)). In the second scenario, component $x$ is essential for correctly separating the classes, and two comparable eigenvalues are associated with the two components.}
    \label{fig:multiplot_appendixA}
\end{figure}

We briefly discuss here a simple generalization of the Gaussian dataset analyzed in the main body of the paper. If the data are characterized by a non-isotropic covariance matrix, the distance between the one-dimensional distributions shown in Figure (\ref{fig:input_distributions_gauss}) is no longer a good estimator of the relative relevance of each input component, since the information is generally distributed across components.
As an example, let us consider a simple case of an input vector composed of only two components, where only one shows partial separation between the two classes, when examining each component separately. Therefore, by inspecting the distributions of the two components, in the spirit of what was done for the Gaussian dataset in the main body of the paper, one might be led to think that only the second component is relevant for solving the task (panel (a) of Figure (\ref{fig:multiplot_appendixA})). However, this is true only if the two components are independent (panel (b) of Figure (\ref{fig:multiplot_appendixA})). When the two components are correlated instead (panel (c) of Figure (\ref{fig:multiplot_appendixA})), the model must take both into account to solve the task efficiently. The post-training eigenvalues associated with the two components, shown in the contour plots for both cases, correctly identify these two different situations. Obviously, this is still a very simple (linearly separable) example, and the results can be compared with other relevance measures obtained from alternative methods, such as the coefficients of a LASSO regressor, which lead to analogous results (data not shown).

\section{Nonlinearly separable dataset with spurious features}

\begin{figure}
     \centering
     \vspace{1cm}
    \includegraphics[width=0.6\textwidth]{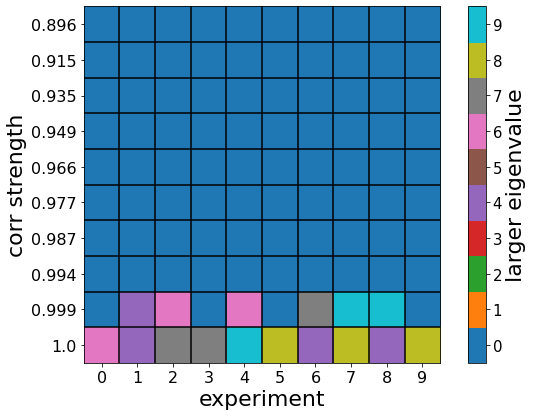}
     \caption{The color represents the larger eigenvalue obtained in each of the 10 different experiments conducted, across 10 different correlation strengths.}
     \label{fig:fraction_appendixB}
\end{figure}

In this appendix, we evaluate the performance of the proposed method on a synthetic dataset that includes spurious input components. A classification task with two distinct classes is simulated. Operatively, each input vector in the dataset consists of ten components: only the first component is used to define the class label, while the others are merely correlated with the first one and are not directly involved in determining the class. Therefore, the underlying problem is essentially a one-dimensional classification task.\\
To work in a setting more suitable for neural networks, the generated classes are not linearly separable. Specifically, we randomly generate the elements of the first component from a uniform distribution between 0 and 1, and we divide this interval into four equal subintervals, assigning each alternately to one of the two classes. Then, we generate the remaining components to be proportional to the first one. Formally:

\begin{equation} 
    \begin{split}
        x_0^d &\in [0,1],\\
        x_i^d &= x_0\alpha_i  \beta^d +\epsilon_i +\eta^d,\\ 
        y^d &=
        \begin{cases} 
            0 \;\; \text{if} \;\; x_0^d \in [0, 0.25) \cup [0.5,0.75)\\
            1 \;\; \text{if} \;\; x_0^d \in [0.25, 0.5) \cup [0.75,1)
        \end{cases},
    \end{split}
\end{equation}

where the index $i$ refers to the components and the index $d$ refers to the data samples. The parameters $\alpha_i$ and $\epsilon_i$ are random values that apply a linear transformation to the first component, while $\beta^d$ and $\eta^d$ are additional random values that may vary across data samples. If $\beta^d = 1$ and $\eta^d = 0$ for all samples, the nine additional components are perfectly correlated with the first one, and the problem can be equivalently represented in the domain of any component. Conversely, if these two parameters vary among the data, the correlation between the additional components and the first one becomes weaker or stronger depending on their variability.\\
We define the correlation strength $s$ between the first component and the others as the average value of the off-diagonal elements of the covariance matrix of the data. The proposed method was tested on different realizations of the dataset while exploring a range of possible correlation strengths.\\
First of all, the method always identifies only one component as relevant, since the information contained in the other components is redundant and the regularization term penalizes solutions that rely on multiple inputs. Moreover, Figure \ref{fig:fraction_appendixB} reports the larger eigenvalue obtained from 10 different experiments and 10 distinct correlation strengths. Even at very high correlation strengths, the first component is correctly identified as the relevant feature. It is only when $s$ approaches 1 (perfect correlation) that the method randomly selects one of the equivalent components to solve the task.

\section{Methods comparison on non-isotropic Gaussian dataset}
\label{apdx:method_comaprison_C}

\begin{figure}
     \centering
     \vspace{1cm}
    \includegraphics[width=0.9\textwidth]{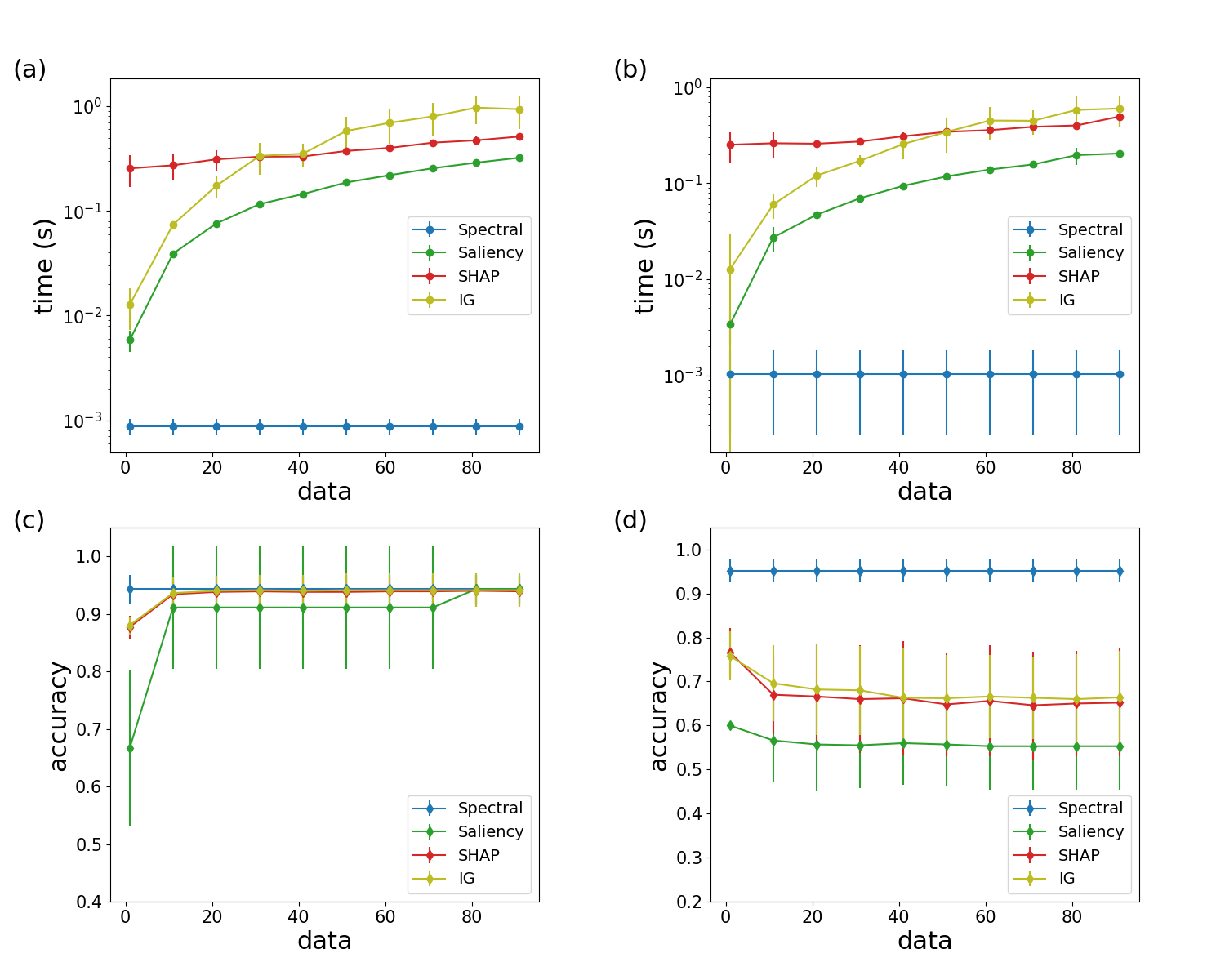}
     \caption{Panels (a) and (b) show the computation time required to estimate the global relevance of input components for the Gaussian dataset using the four analyzed methods: spectral method, saliency map, SHAP, and integrated gradients. Panels (c) and (d) show the accuracy of the same methods in identifying the relevant components.The results in panels (a) and (c) correspond to the case in which all methods are evaluated on the same spectral neural network (trained with an $L_1$ regularization on eigenvalues and eigenvectors). Panels (b) and (d) instead refer to the case where the standard methods (saliency map, SHAP, and integrated gradients) are evaluated on a multilayer perceptron (MLP) with the same architecture as the spectral network. During MLP training, an $L_1$ regularization term is applied to the weights of the first layer to promote sparse solutions. }
     \label{fig:multiplot_appendixC}
\end{figure}

We report here a quantitative comparison between our method and three of the methods described in Section (\ref{sec:comparison}) for a Gaussian dataset. Specifically, a Gaussian dataset consisting of two classes and 100 input features was generated, following the approach presented in Section (\ref{sec:independent_gaussian_results}). The distributions associated with the two classes are partially separated in the first 20 features, which are therefore intrinsically relevant. The next 20 features show no separation between classes but were generated to exhibit a certain degree of correlation with some of the first 20 features and, as explained in Appendix (\ref{appndx:A}), they are consequently relevant for solving the task. Finally, the remaining 60 features are spurious and contain no useful information.\\
We tested the performance of the spectral method, the saliency map, the integrated gradients, and the SHAP method on this class of datasets. In particular, we measured both the accuracy in detecting relevant and non-relevant components and the computation time of each method. The simplicity of this linearly separable dataset allows us to use a LASSO logistic regression model as a benchmark for the number of relevant components. As expected, the LASSO logistic regression identifies the first 40 components as the relevant ones.\\
Since all methods, except for the spectral method, are input-dependent, the global relevance estimate must be obtained by averaging results over multiple examples. Therefore, we report the outcomes obtained for different numbers of samples used to compute the global relevance. Figure (\ref{fig:multiplot_appendixC}) shows the results for 10 independent experiments in two scenarios. In the first one (panels (a) and (c)), all four methods are evaluated on the same model: a spectral neural network trained with an $L_1$ regularization on both eigenvalues and eigenvectors. However, because the $L_1$ regularization on eigenvalues is part of the proposed method itself, we also compare the results obtained with the spectral method with what obtained when the other methods are applied to a standard MLP with an $L_1$ regularization on the weights of the first layer (panels (b) and (d)).\\
In both scenarios, the spectral method proves to be the fastest, with a computation time several orders of magnitude smaller than that of the other methods. The accuracy scores are comparable across all the methods in the first scenario when an adequate number of examples are used to compute the global relevance. Conversely, the accuracy of the other methods when they are tested on standard MLP is considerably lower than that of the spectral method.

\section{Algorithmic complexity of different Methods}
\label{apdx:algo_complexity_D}

We compare here the computational complexity of the methods presented in Section~(\ref{sec:comparison}) with that of the spectral method. In the spectral method, the relevance is computed by multiplying the vector of eigenvalues from the first layer by the norm of the corresponding eigenvector entries. The algorithmic complexity of this operation is $O(n_1 \times n_2)$, where $n_1$ and $n_2$ are the numbers of nodes in the first and second layers, respectively. The algorithmic complexity of the saliency map is on the order of a single forward (or backward) pass \cite{simonyan2013deep,ancona2018towards}, denoted as $C_f$, which is larger than $n_1 \times n_2$ for a multilayer perceptron (MLP). The LRP method exhibits the same complexity~\cite{montavon2018methods}. The Integrated Gradients method introduces an additional cost due to the $m$ steps required to approximate the path integral~\cite{sundararajan2017axiomatic,kapishnikov2021how}. Finally, the computational cost of SHAP is $O(2^{n_1} C_f)$~\cite{lundberg2017unified}.\\
Recall that, except for the spectral method, all the other methods are designed to provide input-dependent explanations. Therefore, to obtain a global relevance estimate for the input components, it is necessary to average the results over $E$ examples. Table~(\ref{tab:complexity}) summarizes the computational complexity of all the methods considered.

\begin{table}[h!]
\centering
\caption{Computational complexity of local and global relevance computation for various explanation methods.}
\begin{tabular}{lcc}
\hline
\textbf{Method} & \textbf{Local Complexity} & \textbf{Global Complexity (E examples)} \\
\hline
Saliency Maps & $O(C_f)$ & $O(E \cdot C_f)$ \\
Layer-wise Relevance Propagation (LRP) & $O(C_f)$ & $O(E \cdot C_f)$ \\
Integrated Gradients & $O(m \cdot C_f)$ & $O(E \cdot m \cdot C_f)$ \\
SHAP (exact) & $O(2^{n_1} \cdot C_f)$ & $O(E \cdot 2^{n_1} \cdot C_f)$ \\
Spectral Method & -  & $O(\cdot n_1 \cdot n_2)$ \\
\hline
\end{tabular}
\label{tab:complexity}
\end{table}

\section{Effect of training set size and regularization strength on method performance}

\begin{figure}
     \centering
     \vspace{1cm}
    \includegraphics[width=0.9\textwidth]{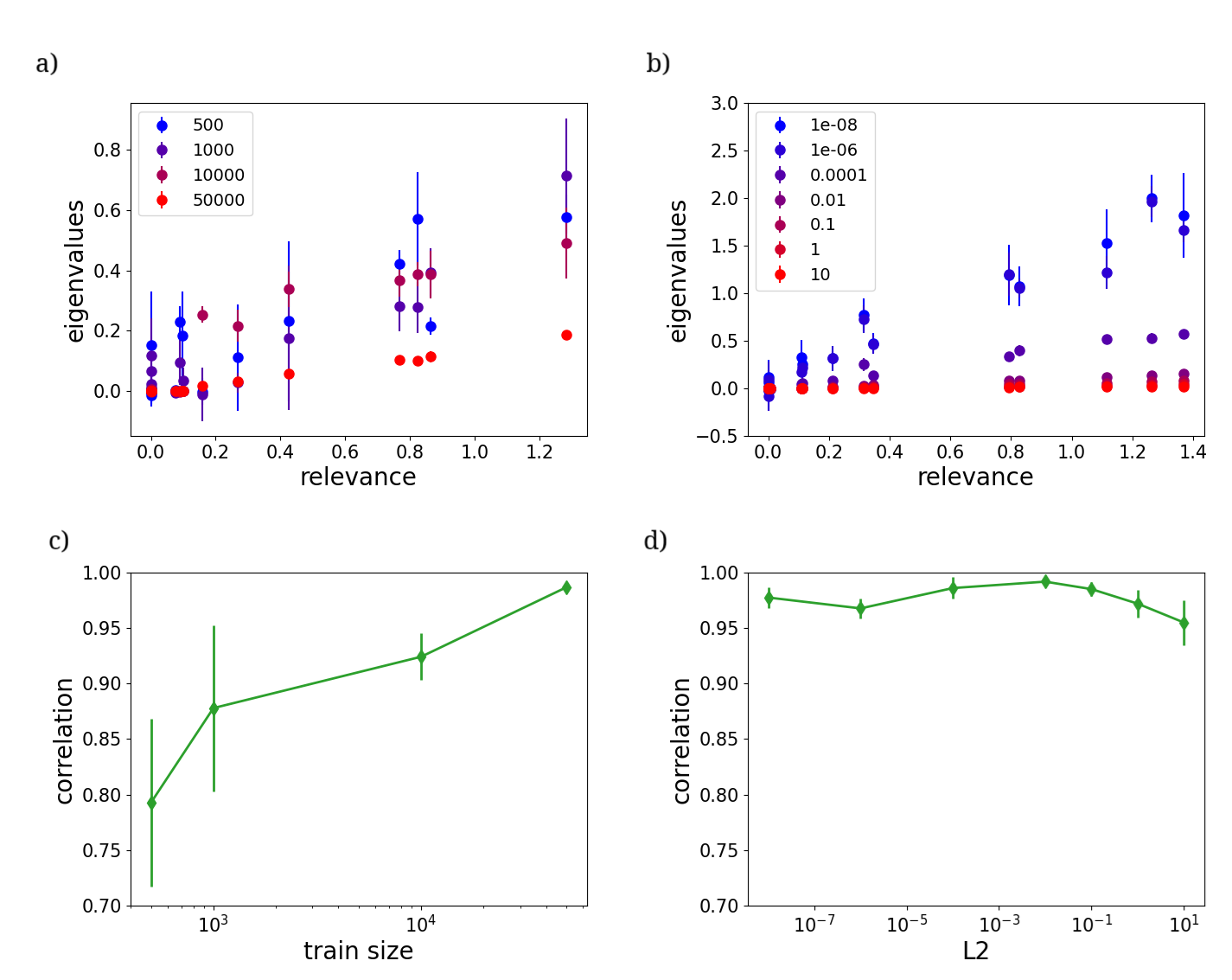}
     \caption{Panels (a) and (b) show the relationship between the eigenvalues and the relevance for different training set sizes and regularization factors, respectively. Panels (c) and (d) show the correlation between the eigenvalues and the relevance scores for the same cases.}
     \label{fig:multiplot_appendixE}
\end{figure}

We analyze here the performance of the proposed method on the first dataset introduced in the manuscript (independent Gaussian dataset) for different training set sizes and various values of the regularization factor applied to the eigenvalues. Figure~(\ref{fig:multiplot_appendixE}) reports the post-training eigenvalues as a function of the relevance defined in Section~(\ref{sec:independent_gaussian_results}) for different training set sizes (panel (a)) and different $L_2$ regularization strengths (panel (b)). We also compute the correlation between the eigenvalues and the relevance for all the considered cases (panels (c) and (d)).\\
The correlation decreases when the training set becomes sufficiently small, likely due to sampling fluctuations during dataset generation, which can lead to situations where the actual relevance of the components differs from the theoretical one. This interpretation is supported by the large standard deviations observed in this regime.\\
Regarding the effect of the regularization factor, the correlation remains high for all tested values, indicating that the eigenvalues are organized consistently with the relevance of the corresponding nodes and that their relative ordering is preserved. However, the absolute magnitudes of the eigenvalues are influenced by the strength of the regularization, as can be observed in panel (b).

\end{document}